%% file: main.tex
\documentclass[10pt,journal,compsoc]{IEEEtran}

\ifCLASSOPTIONcompsoc
\usepackage[nocompress]{cite}
\else
\usepackage{cite}
\fi

\ifCLASSINFOpdf
\else
\fi

\usepackage[pdftex]{graphicx}

\usepackage{gensymb}
\usepackage[table]{xcolor}
\definecolor{Gray}{gray}{0.9}
\usepackage{amsmath,amssymb,amsfonts}
\usepackage{graphicx}
\usepackage{textcomp}
\usepackage{tikz}
\usetikzlibrary{bayesnet}

\usepackage{mathrsfs}
\usepackage[colorlinks,linkcolor=red,citecolor=blue]{hyperref}
\usepackage{amsmath}
\usepackage{amsfonts,subfigure}
\usepackage{amssymb}
\usepackage{multirow}
\usepackage{bm}
\usepackage{sansmath}
\usepackage{booktabs}
\usepackage[ruled]{algorithm2e}

\usepackage{graphicx}
\usepackage{amsthm}

\usepackage{tabularx, makecell, multirow} 
\graphicspath{{figure/}}

\newtheorem{definition}{Definition}[section]

\usepackage{bm}
\usepackage{color,soul}

\usepackage{tikz}

\usepackage{multirow}
\usepackage{adjustbox}
\usepackage{environ}
\usepackage{setspace} 

 \usepackage{epigraph}
\usepackage{enumitem}

\newcommand{\eqnref}[1]{Eq. (\ref{#1})}

\newcommand{\tableref}[1]{Table~\ref{#1}}
  
\newcommand{\figref}[1]{Figure~\ref{#1}} 
\graphicspath{{image/}}

\usepackage{makecell}
\definecolor{Gray}{gray}{0.9}

\begin{document}
	
	\title{Robust Traffic Forecasting  against \\ Spatial Shift over Years }
	
	%
	%
	%
	%
	

	\author{Hongjun~Wang,
	Jiyuan~Chen,
	Tong~Pan,
	Zheng~Dong,\\
	Lingyu~Zhang,
	Renhe~Jiang,
	and
	Xuan Song
	\IEEEcompsocitemizethanks{
		\IEEEcompsocthanksitem Hongjun Wang, Jiyuan Chen,  Tong Pan,  Zheng Dong  and Lingyu Zhang are with (1) SUSTech-UTokyo Joint Research Center on Super Smart City, Department of Computer Science and Engineering
		(2) Research Institute of Trustworthy Autonomous Systems, Southern University of Science and Technology (SUSTech), Shenzhen, China.
		E-mail: {wanghj2020,11811810}@mail.sustech.edu.cn, pant@sustech.edu.cn,  zhengdong00@outlook.com, and zhanglingyu@didiglobal.com. 
		\IEEEcompsocthanksitem Xuan Song is with (1) School of Artificial Intelligence, Jilin University  (2) Research Institute of Trustworthy Autonomous Systems, Southern University of Science and Technology (SUSTech), Shenzhen, China. Email: songxuan@jlu.edu.cn.
		\IEEEcompsocthanksitem R. Jiang is with Center for Spatial Information Science, University of
		Tokyo, Tokyo, Japan. Email: jiangrh@csis.u-tokyo.ac.jp.
		\IEEEcompsocthanksitem Corresponding to  Xuan Song;
	}
}

	%
	%

\markboth{Journal of \LaTeX\ Class Files,~Vol.~XX, No.~X, August~201X}%
{Shell \MakeLowercase{\textit{et al.}}: Bare Demo of IEEEtran.cls for Computer Society Journals}
%

\IEEEtitleabstractindextext{%
	\begin{abstract}
Recent advancements in Spatiotemporal Graph Neural Networks (ST-GNNs) and Transformers have demonstrated promising potential for traffic forecasting by effectively capturing both temporal and spatial correlations. The generalization ability of spatiotemporal models has received considerable attention in recent scholarly discourse.
However, no substantive datasets specifically addressing traffic out-of-distribution (OOD) scenarios have been proposed. Existing ST-OOD methods are either constrained to testing on extant data or necessitate manual modifications to the dataset. Consequently, the generalization capacity of current spatiotemporal models in OOD scenarios remains largely underexplored. In this paper, we investigate state-of-the-art models using newly proposed traffic OOD benchmarks and, surprisingly, find that these models experience a significant decline in performance. Through meticulous analysis, we attribute this decline to the models' inability to adapt to previously unobserved spatial relationships.
To address this challenge, we propose a novel Mixture of Experts (MoE) framework, which learns a set of graph generators (i.e., graphons) during training and adaptively combines them to generate new graphs based on novel environmental conditions to handle spatial distribution shifts during testing. We further extend this concept to the Transformer architecture, achieving substantial improvements. Our method is both parsimonious and efficacious, and can be seamlessly integrated into any spatiotemporal model, outperforming current state-of-the-art approaches in addressing spatial dynamics. \textcolor{magenta}{Codes are available at \textcolor{black}{\href{https://github.com/Dreamzz5/ST-Expert}{GitHub}}}. 
	\end{abstract}
	
	}

\maketitle
\section{Introduction}
\label{sec:intr}
\IEEEPARstart{T}{raffic} forecasting \cite{li2017diffusion,wu2019graph,yan2018spatial,jain2016structural} has emerged as a powerful technique for modeling dynamic systems, gaining prominence with the advancements in graph neural networks. It effectively captures the inter-dependencies between connected nodes, which has been successfully applied in a wide range of complex system problems.

Despite this, it has come to our attention that current ST Graph Neural Networks (ST-GNNs) are primarily evaluated based on short-term training and testing distributions, typically spanning only three months \cite{guo2019attention,li2017diffusion,yu2017spatio}. Within such short period of time, basically there will be no spatial or temporal pattern shifts. However, cities are constantly evolving and developing.
\textit{ For example, imagine a new highway is built in a city or a popular shopping mall opens in a different area. These changes can lead to shifts in traffic flow, creating new dependencies between previously unrelated regions or altering existing traffic patterns \cite{zhang2020curb,zhang2019trafficgan,zhang2019gcgan}.}
 In long-term scenarios, say one year, such possibility can greatly increase. Therefore, we think the current way of measuring ST-GNNs' performances overlooks the intricate nature of dynamic ST-dependencies, and can not reflect the true ability of ST-GNNs in handling long-term scenarios with unseen spatial/temporal patterns.

\begin{table}[t]
	\centering
	\renewcommand{\arraystretch}{1.1}
	\caption{ Information of the shifted datasets used in this paper}\label{tab:detail}
	\resizebox{0.5\textwidth}{!}{\begin{tabular}{ccccc}
			\hline    Datasets & Train Year & Train Peroids & Test Year&  Test Peroids  \\
			\hline
			PEMS03 & 2018 & 09/01 - 11/12 & 2018/2019 & 11/13 - 11/30  \\
			PEMS04 & 2018 & 01/01 - 02/16 &2018/2019&  02/17 - 02/28 \\
			PEMS07 & 2017 & 05/01 - 08/06  &2017/2018& 08/07 - 08/31 \\
			PEMS08 & 2016 & 07/01 - 08/19 &2016/2017&   08/20 - 08/31 \\
			SpeedNYC & 2019 & 03/01 - 05/12 &2020-2022&   05/13 - 08/31 \\
			\hline
	\end{tabular}}
\end{table}

\begin{figure*}[t]
	\centering
	\subfigure[Performance comparison on PEMS03]{\includegraphics[width=.48\linewidth]{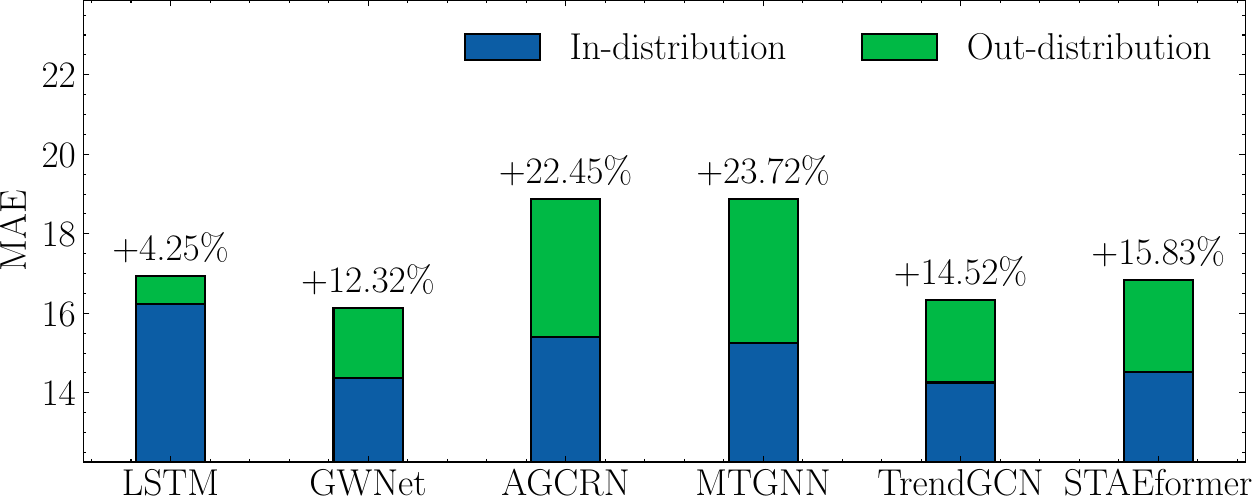} \label{fig:compare_1}}
	\subfigure[Performance comparison on PEMS04]{\includegraphics[width=.48\linewidth]{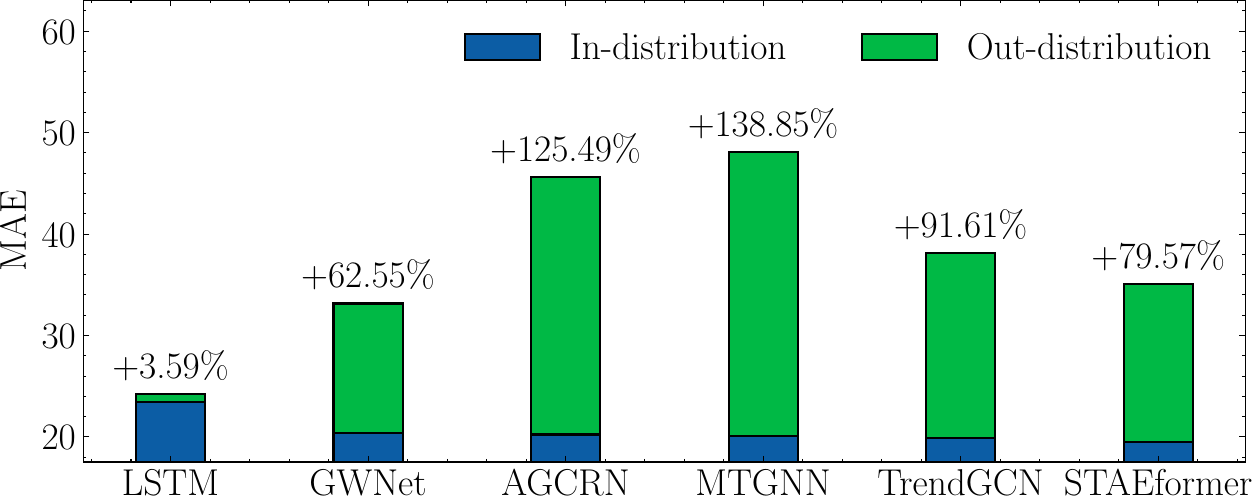}  \label{fig:compare_2}}
	\subfigure[Spatiotemporal discrepancy on PEMS03]{\includegraphics[width=0.48\linewidth]{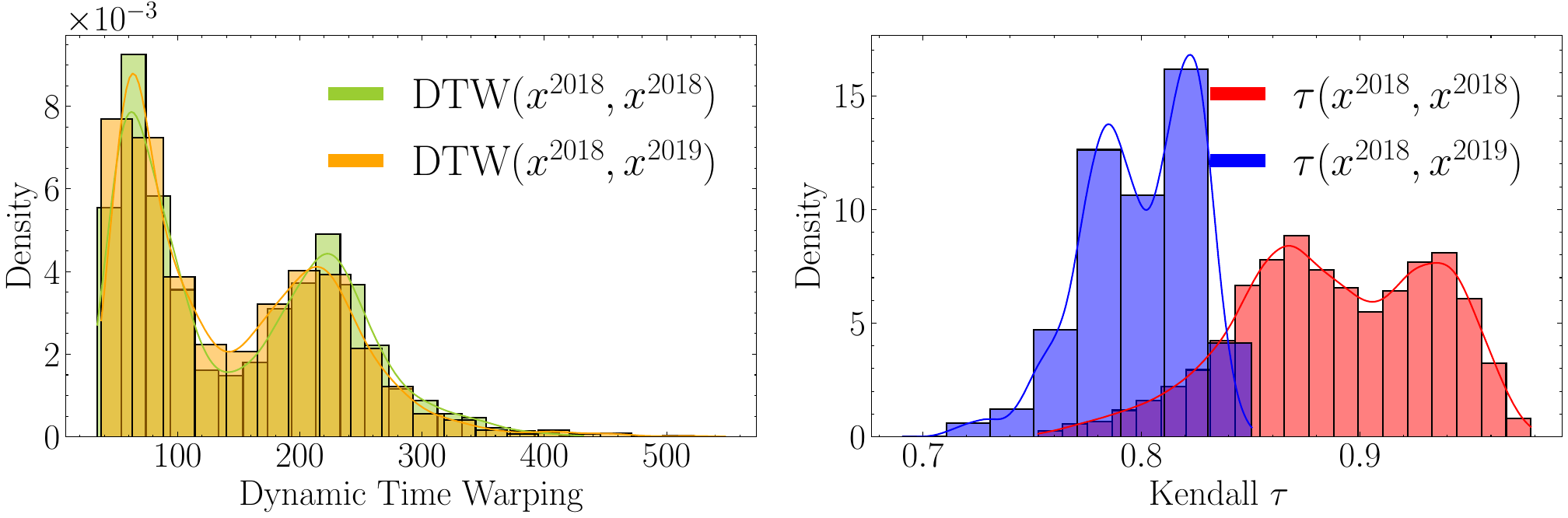} \label{fig:kendall_03}}
	\subfigure[Spatiotemporal discrepancy on PEMS04]{\includegraphics[width=0.48\linewidth]{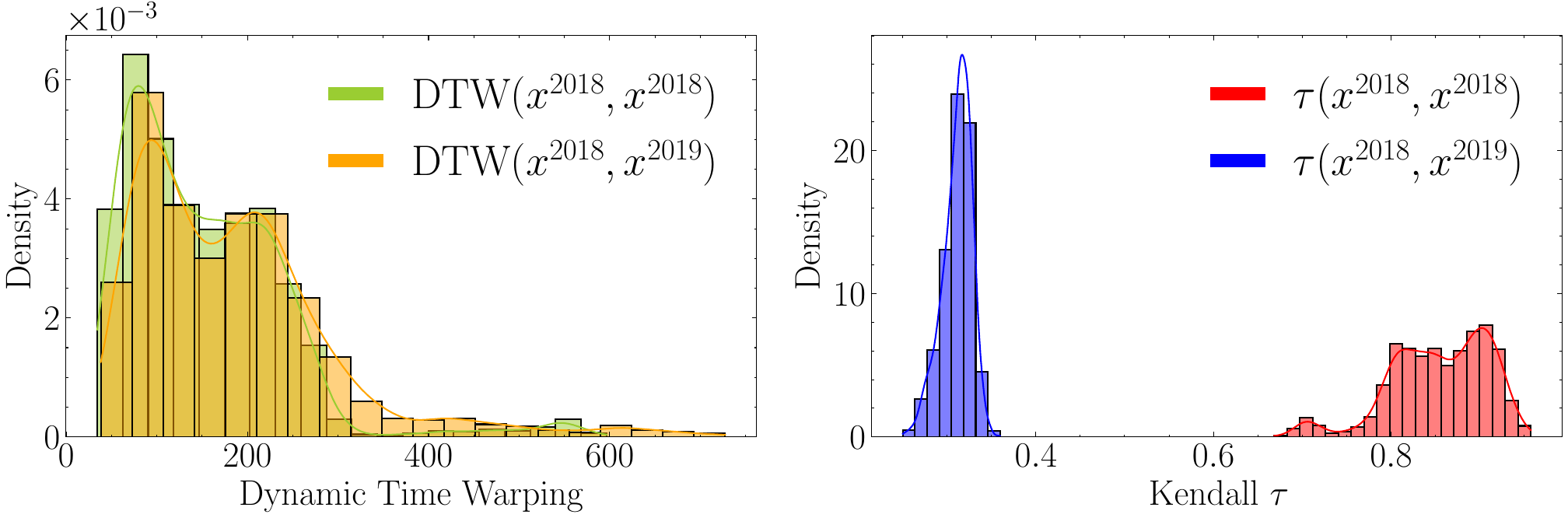} \label{fig:kendall_04}}
	
	\vspace{-5pt}
	\caption{In (a) and (b), we compare the test performance of mainstream ST-GNNs: GWNet~\cite{wu2019graph}, AGCRN~\cite{bai2020adaptive}, MTGNN~\cite{wu2020connecting}, TrendGCN~\cite{jiang2023enhancing} and STAEformer~\cite{liu2023spatio} on in- and out-of-distribution, respectively. The statistical results for PEMS03 and PEMS04 are shown in (c) and (d), respectively. For each dataset, we illustrate the distribution of Kendall’s $\tau$ \cite{kendall1938new} and DTW \cite{muller2007dynamic}, which present the similarity of graph relations and temporal distributions. 
	}\label{fig:motivation}
\end{figure*}

Recent pioneering efforts \cite{xia2024deciphering,zhou2023maintaining,wang2024stone} have made significant progress in addressing the challenge of out-of-distribution scenarios in traffic forecasting. However, these methods are either tested on data from recent weeks \cite{xia2024deciphering} or involve modifications to the dataset \cite{zhou2023maintaining,wang2024stone}, such as manually masking certain nodes to create spatial shifts. They never make a preliminary experiment in the data level to see whether the shift they claim truly exist.

To address this issue, we  takes the first step to examine the existence of such shift from benchmark datasets with novel experiments and propose four OOD traffic  benchmarks: PEMS03-2019, PEMS04-2019, PEMS07-2018, and PEMS08-2017, derived from the California Department of Transportation Performance Measurement System (PEMS)\footnote{\url{https://pems.dot.ca.gov/}} \cite{chen2001freeway}. These benchmarks maintain \textbf{identical sensors} while capturing data from \textbf{different years}, aligning with existing standards \cite{song2020spatial}. For example, PEMS03 in the previous benchmark recorded traffic flow from September to November 2018, while PEMS03-2019 refers to traffic signals from September to November 2019.  To comprehensively evaluate the model's ability to generalize to unseen data, we leveraged real-time traffic information obtained from the New York City Department of Transportation's Traffic Management Center (TMC))\footnote{\url{https://www.nyc.gov/html/dot/html/motorist/atis.shtml}}. Specifically, the model was trained on historical Speed data encompassing the period from March to May 2019. Subsequently, its performance was assessed using a multi-year dataset spanning 2020 to 2022, which served as unseen data for the model. Dataset details are presented in  \tableref{tab:detail}. Date consistency was maintained to control for seasonal effects. 
 
 Models were trained on original benchmarks and tested on both in-distribution and OOD data. Performance results for prominent ST-models are summarized in \figref{fig:motivation}. \textit{(Additional baseline performances are provided in  \tableref{tab:forecast_result}.)}  However, we observed significant performance degradation in ST-models when tested over extended time spans. We hypothesize that this deterioration stems from shifting graph relations, supported by our ablation study using an LSTM (\figref{fig:compare_1} and \ref{fig:compare_2}), which considers only temporal dependencies and maintains consistent performance across both in-distribution and OOD data.

Our findings align with the observations in \cite{wang2021exploring}, which suggest that temporal knowledge tends to be more generalizable than spatial knowledge. Spatial knowledge requires more careful selection, as blindly increasing spatial information may reduce both the effectiveness and efficiency of the model. In traffic scenarios, we believe that the factors limiting temporal OOD  may be road-related, such as traffic flow being influenced by the number of lanes \cite{liu2024largest} and traffic speed being constrained by speed limits \cite{chen2001freeway}.


We delve deeper into the distinction between in- and out-of-distribution scenarios. To assess the similarity of graph relations, we employ Kendall's $\tau$ coefficient \cite{kendall1938new}, a statistic that measures rank correlation between variables, ranging from -1 to 1, with higher values indicating stronger correlation \cite{edwards2023graphing}. A higher $\tau$ value indicates greater consistency in the ordering of variables, with 1 representing perfect positive correlation, -1 perfect negative correlation, and 0 no correlation. Formally, Kendall's $\tau$ for node $v$ is defined as:

\[
\tau_v=\frac{2}{n(n-1)} \sum_{u \in N(v)} \sum_{i < j} \ \operatorname{sign}\left(x^{(i)}_v - x^{(i)}_u\right) \operatorname{sign}\left(y^{(j)}_v - y^{(j)}_u\right),
\]

where $x_v$ and $y_v$ indicate the signal of node $v$ in the same year or across different years, $N(v)=\{u \in \mathcal{V} \mid(v, u) \in \mathcal{E}\}$ represents the neighbors of node $v$, $n = |N(v)|$ indicates the number of neighbors for node $v$, and $\mathcal{E}$ denotes the set of edges in the graph, and $\operatorname{sign}$  function returns +1 if $x_v > x_u$, -1 if $x_v < x_u$, and 0 if $x_v = x_u$. 

We also introduce the Dynamic Time Warping (DTW) \cite{muller2007dynamic}, a metric for measuring the similarity between time series, where a smaller value indicates greater similarity. DTW values are always non-negative, with 0 indicating identical sequences and larger values representing greater dissimilarity between time series.
The statistical results are presented in \figref{fig:kendall_03} and \ref{fig:kendall_04}, where the left figure illustrates the DTW distance distribution across all sensors, and the right part displays the coefficient of Kendall's $\tau$  between in and out of distribution. 
Notably, the DTW distance distribution shows a high consistency when compared with the Kendall's $\tau$ distribution, which explain why LSTM maintains overwhelming performance in long-term traffic forecasting.

To address the spatial dynamic nature of traffic datasets, we draw inspiration from the successful implementation of the mixture of experts model \cite{jacobs1991adaptive} in domain generalization \cite{li2022sparse}. 
We proposes an expert graphon layer, where each graphon represents a graph generator. During training, we aim for each expert to learn as much as possible environment knowledge, and we employ episodic eraining \cite{li2019episodic} to expose the model to OOD environments, which enables the model to adaptively combine individual graphs during testing to generate new graphs suited to novel environments. The expert graphon layer can be seamlessly integrated with any ST-GNN that includes learnable graph modules, and it can be easily extended to Transformer architectures  to handle spatial drift scenarios. We achieve robust performance on both existing and newly proposed traffic benchmarks, serving as a gentle remedy for stable traffic prediction.

\begin{itemize}
	\item[$\bullet$] We propose a set of novel traffic out-of-distribution  benchmarks and perform a thorough evaluation of state-of-the-art spatiotemporal models. Our analysis reveals a substantial performance degradation in OOD scenarios, primarily due to the models' inability to adapt to unobserved spatial relationships.
	
	\item[$\bullet$] We introduce a novel Mixture of Experts  framework, which learns a set of graph generators (graphons) during training. These graphons are adaptively combined to synthesize new graphs under novel environmental conditions, thereby effectively addressing spatial distribution shifts during testing.
	
	\item[$\bullet$] We extend our MoE framework to the Transformer architecture, yielding significant performance improvements. The proposed approach is both parsimonious and effective, and can be effortlessly integrated into any spatiotemporal model. Our method consistently outperforms existing state-of-the-art approaches in capturing spatial dynamics.
\end{itemize}

\begin{figure*}[t]
	\centering
	\includegraphics[width=0.85\linewidth]{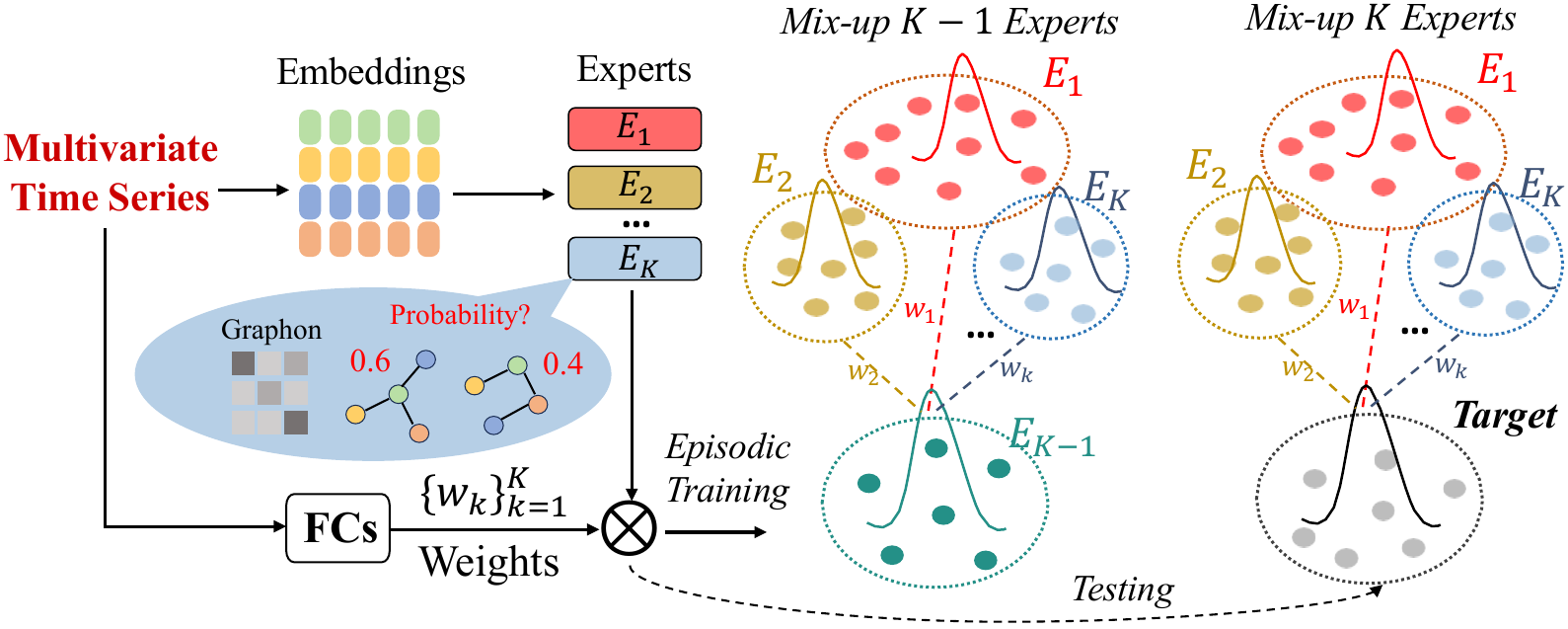}
	\caption{The schematic of targeted graphons generation in training and testing period. }\label{fig:pipeline}
\end{figure*}

\section{Related Work}
\subsection{Traffic Forecasting}
The primary challenge in traffic forecasting is capturing both spatial and temporal dependencies from dynamic inputs against a fixed graph structure. Recent solutions primarily incorporate graph neural networks to model the complex spatial relationships in traffic networks \cite{kipf2017semi,li2018diffusion,yu2018spatio}. Early approaches like DCRNN \cite{li2018diffusion} and STGCN \cite{yu2018spatio} pioneered the integration of graph convolution with recurrent or convolutional structures for traffic forecasting.
Subsequent works have explored various techniques to enhance spatio-temporal representation learning. For instance, Graph WaveNet \cite{wu2019graph} introduced a self-adaptive adjacency matrix to capture hidden spatial dependencies. ASTGCN \cite{guo2019attention} and STSGCN \cite{song2020spatial} employed attention mechanisms to dynamically capture spatial and temporal correlations. More recent models like STFGNN \cite{li2021spatial} and D$^{2}$STGNN \cite{shao2022decoupled} have proposed novel architectures for joint spatial-temporal dependency modeling.
However, most existing research primarily evaluates ST-GNNs within short timeframes, potentially overlooking long-term dynamics in traffic data. Recent studies \cite{xia2024deciphering,zhou2023maintaining,wang2024stone} have attempted to address this limitation by exploring model generalization from a temporal perspective. For example, \cite{xia2024deciphering} proposed a causal framework to analyze the long-term effectiveness of ST-GNNs, while \cite{zhou2023maintaining} introduced a maintain-and-attain mechanism to enhance long-range forecasting.
Nevertheless, upon thorough analysis of real traffic data from sources like PEMS, we find that addressing spatial shift presents an even more formidable challenge. This is due to the inherent complexity and heterogeneity of spatial relationships in traffic networks, which can vary significantly across different regions and time periods. Our work aims to specifically tackle this spatial dynamic challenge, complementing existing temporal-focused approaches to provide a more comprehensive solution for traffic forecasting.

\subsection{OOD in Spatio-Temporal Analysis}
OOD generalization in graph-structured data has emerged as a critical challenge in recent years. Li et al. introduced OOD-GNN \cite{li2022ood}, which enhances GNNs ability to generalize to unseen graph structures by learning to decorrelate causal and non-causal features. Similarly, Park et al. proposed MAGNA \cite{park2021metropolis}, leveraging a Metropolis-Hastings data augmentation technique to improve GNN performance on OOD graphs. Wu et al. \cite{wu2022handling} adopted an invariance-based approach, focusing on learning invariant rationales to manage distribution shifts on static graphs. However, these methods are primarily tailored to static graphs and overlook the temporal dynamics often present in real-world evolving networks.
In contrast, time series OOD detection and generalization have gained prominence due to the non-stationarity of many temporal processes. Wu et al. introduced DIVERSIFY \cite{wu2022diversify}, a framework that disentangles seasonal-trend representations for time series OOD generalization. Yang et al. \cite{yang2022towards} proposed a causal approach for OOD sequential event prediction, addressing evolving temporal patterns, while Du et al. developed AdaRNN \cite{du2021adarnn}, an adaptive method for forecasting OOD time series by dynamically adjusting to varying contexts. Despite their successes in managing temporal shifts, these methods fall short of explicitly modeling spatial relationships, limiting their applicability in scenarios where both spatial and temporal distributions change.
The integration of spatial and temporal dimensions in OOD scenarios poses additional challenges, particularly in fields like traffic prediction and climate modeling. Recent efforts have begun addressing this issue. Xia et al. \cite{xia2023deciphering} developed CaST, a causal framework for transferring invariant spatio-temporal relations to OOD settings, while Zhou et al. proposed CauSTG \cite{zhou2023maintaining}, aimed at capturing invariant relations in spatio-temporal learning. Additionally, Hu et al. \cite{hu2023graph} introduced a graph neural process for spatio-temporal extrapolation, partially addressing OOD generalization. Wang et al. \cite{wang2024stone} proposed STONE, a novel framework that generates invariant spatiotemporal representations for effective generalization to unknown environments through semantic graph learning, graph intervention mechanisms, and an Explore-to-Extrapolate loss.

\section{Problem Statements}
In this paper, we define a graph as \(\mathcal{G}=(\mathcal{V}, \mathcal{E}, A)\), where \(\mathcal{V}\) represents the set of nodes, \(\mathcal{E} \subseteq \mathcal{V} \times \mathcal{V}\) defines the edges, and \(A\) is the adjacency matrix associated with the graph \(\mathcal{G}\). Additionally, at each time step \(t\), the graph possesses a dynamic feature matrix \(\mathbf{X}_t\) within the real-number space \(\mathbb{R}^{|\mathcal{V}| \times \mathcal{C}}\), with \(\mathcal{C}\) indicating the dimensionality of the node features. 
The task of traffic forecasting involves developing and training a neural network model \(g\). The functional relationship for this predictive modeling is formulated as: $g: \left[X_t, A\right] \mapsto Y_t$, 
where \(X_t = \mathbf{X}_{(t-l_1): t}\) and \(Y_t = \mathbf{X}_{(t+1):(t+l_2)}\), with \(l_1\) and \(l_2\) representing the lengths of the input and output sequences, respectively.


Current  spatiotemporal OOD methods \cite{xia2024deciphering,zhou2023maintaining,wang2024stone} commonly assume a invariant graph relationship $\mathcal{R}$ in dynamic feature matrices $\mathbf{X}_t$, with the expectation that trained models will perform well on future unseen scenario. In fact, we posit that the notion of spatiotemporal invariance is inherently ill-defined. For instance, predicting when future traffic condition might occur due to the construction or development of new commercial centers is inherently uncertain.   Our work explores a more practical scenario where the training and testing graph relationships may differ, and the training graph relationship also evolves over time. Following \cite{peters2016causal}, we define a graph environment as the joint distribution $P_{XAY}$ across $X_t \times A_t \times Y_t$, denoted by $\Omega$. For each environment, the graph relation $\mathcal{R}_t=\left(X_t, A_t, Y_t\right)$ exists, but feature and outcome distributions can vary ($P_{X A Y}^e \neq P_{X A Y}^{e^{\prime}}$ for different $e, e^{\prime} \in \Omega$). Thus, the learning goal of this paper is to construct a set of experts and each expert learn a specific environment $e$ for stable predictions across various environments, despite selection biases.

\begin{figure*}[t]
	\centering
	\includegraphics[width=0.8\linewidth]{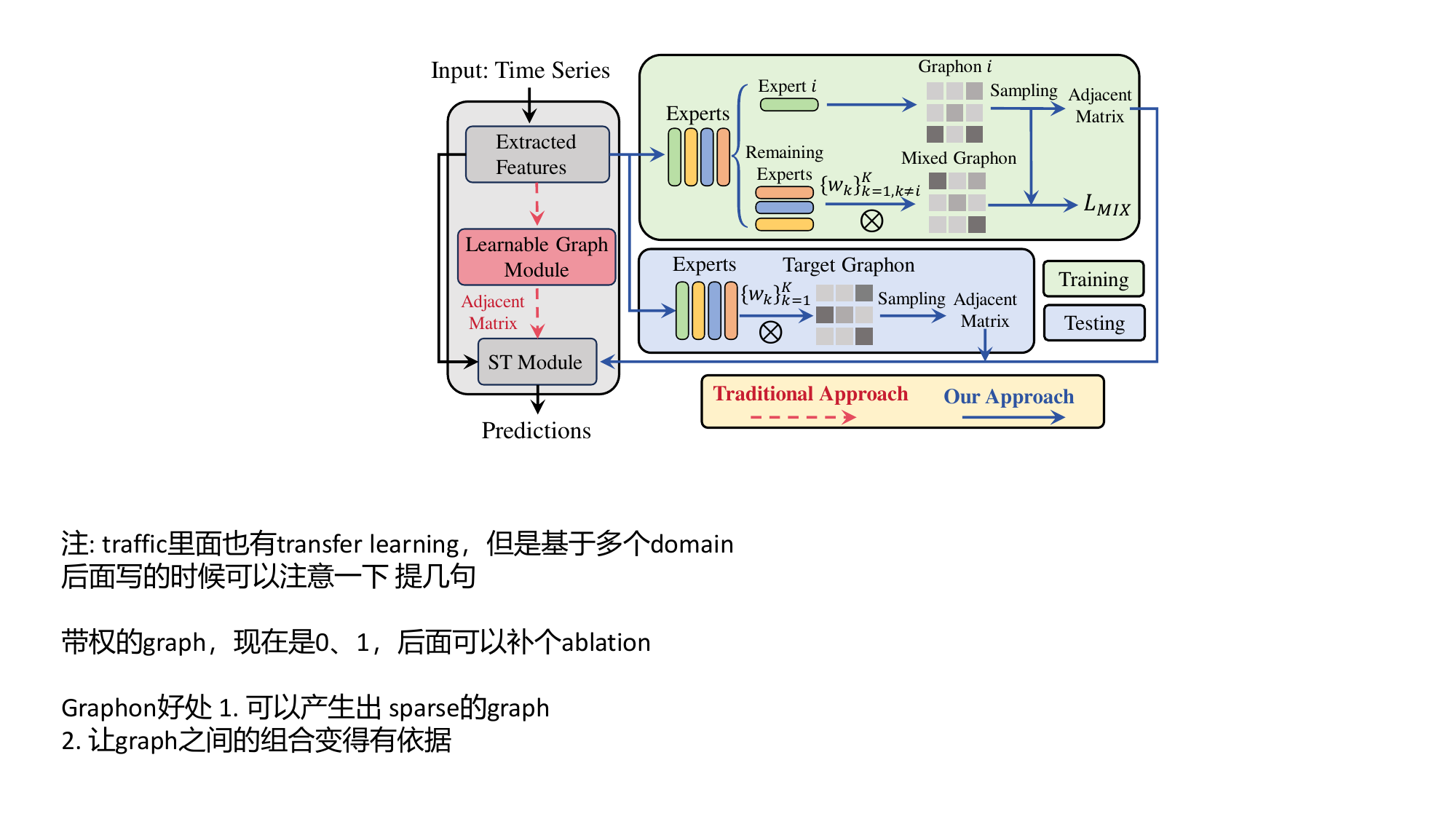}
	\caption{Our method integrates seamlessly with conventional ST-GNNs by incorporating a learnable expert graphons layer. The left (grey) portion of the figure represents the traditional ST-GNNs pipeline, which we enhance by replacing the standard learnable graph construction module with our expert graphons framework. Both approaches serve the core function of preparing the adjacency matrix for the spatiotemporal module, ensuring compatibility and easy integration.}\label{fig:framework}
\end{figure*}
\section{Methodology}
As discussed, our objective is to develop a spatiotemporal model capable of learning diverse graph generators (graphons) that adapt to specific environmental conditions in unseen data. To achieve this, we seek to identify periods with similar graph relations \(\mathcal{R}\), enabling each expert to learn distinct patterns independently. This approach builds on \cite{zhou2023maintaining}, which emphasizes dataset diversity for robust model training, but relies on manual heuristics for partitioning, potentially leading to suboptimal outcomes. We address this by introducing the Maximum Spatiotemporal Graph Division problem, which partitions the dataset into \(K\) distinct time intervals while maximizing dissimilarity between them. Formally, the problem is defined as follows:
\begin{definition}[Maximum Spatiotemporal Graph Division (MSGD)]
	Let \(\mathcal{T}\) be a spatiotemporal dataset that can be decomposed into \(K\) distinct time intervals, denoted as \(\mathcal{T} = \{\mathcal{T}_1, \mathcal{T}_2, \ldots, \mathcal{T}_K\}\). Each interval \(\mathcal{T}_k\) consists of pairs \(\{X_t, Y_t\}^{t_{k+1}}_{t=t_k}\), where \([t_k, t_{k+1})\) denotes the continuous time interval range, such as morning peak and evening peak. The MSGD is characterized by a scenario where, within each individual period \([t_k, t_{k+1})\), all data segments conform to the same graph relationship \(\mathcal{R}_k\). Conversely, for any two distinct time periods \(i\) and \(j\) where \(1 \leq i, j \leq K\) and \(i \neq j\), the graph relationships differ, i.e., \(\mathcal{R}_{\mathcal{T}_j} \neq \mathcal{R}_{\mathcal{T}_j}\). \label{def}
\end{definition}
Based on the principle of maximum entropy \cite{jaynes1982rationale} and Def. \ref{def}, we accomplishes this by addressing an optimization problem with the following objective function:
\begin{equation}
	\label{equ:obj}
	\begin{split}
		\max_{0 < K \le |\mathcal{T}|} \, {\max_{t_1, \cdots, t_K}   \frac{1}{K}\sum_{1 \le i \ne j \le K} d(\mathcal{T}_i, \mathcal{T}_j)}\\
		\mathrm{s.t.} \,\, \forall  \, i, \,   \alpha_1 < |\mathcal{T}_i| < \alpha_2; \, \sum_{i} |\mathcal{T}_i| = T,
	\end{split}
\end{equation}
where  distance metric \(d(\cdot,\cdot)\) is chosen as Kendall’s $\tau$ to measure the graph relation across different periods of graph signals, \(\alpha_1\) and \(\alpha_2\) are predefined parameters representing the minimum and maximum interval ranges, respectively.
\(|\mathcal{T}_k| = t_{k+1} - t_k\) denotes the length of each interval, and \(T\) denotes the total time interval for one day.

The optimization problem in ~\eqnref{equ:obj} aims to maximize the average distribution dissimilarity among periods, which is achieved by searching for an optimal \(K\) and the corresponding periods, ensuring that each period's distribution is as diverse as possible.  
While no prior assumptions are made about this problem, it is reasonable to assume that the graph relations at specific time period across different day are highly consistent. For example,  during the morning peak on weekdays, we presume that the graph relationships remain consistent across different days, which allows for more flexible and general expert graphons. 
In practice, solving the splitting optimization problem in~\eqnref{equ:obj} can be computationally challenging and might not have a closed-form solution. Instead, we here try to 
utilize dynamic programming (DP) \cite{ross2014introduction} to 
select an appropriate splitting number $K$ and corresponding index $\mathcal{T}$.  



\noindent\textbf{Graphons Generator.} In this paper, graphon is formulated as probability matrix, denoted as \(\mathcal{P}\), where \(\mathcal{P}(i, j)\) indicates the likelihood of an edge existing between nodes \(i\) and \(j\). The graphs within one
expert are produced under the same generator (i.e., graphon).
When handle data that falls outside the training distribution,  mix-up process generates optimal graph relations tailored to the current input. A visual example of the process to generate the target distribution is shown in \figref{fig:pipeline}.  To generate graphons, we employ a set of learnable expert embedding matrices, and a dynamic  embedding layer to encode the input signal. Formally, let the \(k^{th}\) expert embeddings as \(E_g^{(k)} \in \mathbb{R}^{|\mathcal{V}| \times d}\), and \(E_t \in \mathbb{R}^{|\mathcal{V}| \times d}\) signifies the time series embedding features dynamically generated by the current input signal \(x\) with multilayer perceptron. To construct the graphons matrix, we  calculate the $k^{th}$ graphons \(\mathcal{P}_k = \sigma\left(E^{(k)}_{g} E_{t}^T\right) \in \mathbb{R}^{|\mathcal{V}| \times |\mathcal{V}|}\), where  \(\sigma\) is the sigmoid function, used to regulate the range of the elements to (0, 1).  

Subsequently, we can sample a graph $G_k$ from graphon $\mathcal{P}_k$ incorporating with the Gumbel softmax \cite{jang2016categorical} for reparameterizing the graph's probability distribution. Moreover, Gumbel softmax can helps in minimizing the impact of less significant values, which could be seen as noise. The formulation of the Sampled Graph is given by:
\begin{align}
	G_k = \sigma\left(\log\left(\frac{\mathcal{P}_k}{1 - \mathcal{P}_k}\right) + \frac{z^1 - z^2}{s}\right),
\end{align}
where \(z^1\) and \(z^2\) are sampled from a $Gumbel(0,1)$ distribution, and \(s\) is a temperature hyperparameter. 

\noindent\textbf{Training Policy.} 
Real-world spatial dependencies are influenced by numerous factors, including irregular urban development, changes in transportation policies, significant events, and more \cite{zhang2020curb,zhang2019trafficgan,zhang2019gcgan}, which presents a challenging problem for developing models that can generalize to novel graph relation different from those encountered during training.
To enhance the ability of each expert to handle out-of-distribution scenarios, we integrate episodic training \cite{li2019episodic}, that trains a single deep network by exposing it to the domain shift scenario. In traffic forecasting, episodic training is to simulate the testing process during training for updating the graphons mixup \cite{han2022g}. An overall pipeline of our training policy is shown in \figref{fig:framework}. For instance, when  $x_i \in \mathcal{T}_i$ is input to the network, domain $\mathcal{T}_i$ is treated as the \textit{unseen target domain} for the other $K-1$ experts $\left\{\mathcal{P}_k \right\}_{k=1,k \neq i}^K$. These $K-1$ experts are combined to generate the mixture's graphons $\mathcal{P}_{mix}$. We aim for $\mathcal{P}_{mix}$ to closely match $\mathcal{P}_{i}$ since the graphon $\mathcal{P}_{i}$ represents the optimal solution for $x_i$. Formally, our training policy is formulated as:
\begin{alignat}{2}
	&\text{Generating Graphon:} &&\quad \mathcal{P}_k = \sigma\left(E^{(k)}_{g} E_{t}^T\right),  \nonumber \\
	& \text{Generating Weight :} &&\quad  \left\{w_k \right\}_{k=1,k \neq i}^K  = f(x_i)\nonumber, \\
	& \text{Graphons Mixup:} &&\quad \mathcal{P}_{mix} = \sum\limits_{k=1,k \neq i}^{K}\frac{e^{w_k}}{\sum\nolimits_{j,j \neq i}e^{w_j}} \mathcal{P}_k,\nonumber\\
	&\text{Feeding into ST-GNNs:}&&\quad \hat{y}_i = g(x_i, G_{k}), \ G_k \sim \mathcal{P}_k, \nonumber 
\end{alignat}
where $w = f(x_i)$ represents the weight of the $k$-th expert, dynamically generated by the input $x_i$. The mixture's graphons $\mathcal{P}_{mix}$ are generated by the weighted average of softmax $w$. Subsequently, the sampled graph $G_k \sim \mathcal{P}_k$ is fed into the ST-GNNs' module $g$ to forecast the future signal $\hat{y}_i$. Our training objective function is defined as:
\begin{align}
	\mathcal{L}_{mix} = \frac{1}{B} \sum_{i=1}^{B}\left(\mathcal{P}_{i}-\mathcal{P}_{mix}^{(i)}\right)^2, \quad \mathcal{L}_{base} = \frac{1}{B} \sum_{i=1}^{B}\left(y_{i}-    \hat{y}_i \right)^2, \label{equ:loss2}
\end{align}
where  $B$ denotes the batch size, $\mathcal{L}_{base}$ serves as the training  objective for the backbone, and $\mathcal{L}_{mix}$ represents the episodic loss designed to strengthen the experts' ability to mix together and handle the out-of-distribution scenarios. It's noteworthy that during the training phase, $\mathcal{L}_{mix}$ is employed to only update the mixup weight $w$.  We interrupt the gradient propagation within each expert graphon $\mathcal{P}_k$ to preserve the independent properties of each expert.

\noindent\textbf{Testing Policy.} In the testing phase, we iteratively create graphons for each sample by
\begin{align}
	\mathcal{P}_{1:K}=\{\sigma\left(E^{(k)}_{g} E_{t}^T\right) : k \in \{1,\cdots, K\}\}. 
\end{align}
Subsequently, we generate mixup weights $w=f(x_{test})$ based on the test signal. Following this, we multiply the weights with the corresponding graphons to produce new graphons. 
It is noteworthy that during the testing period, all graphons participate in the comparison, unlike during training, where the optimal graphon is known. Finally, we sample a new graph and input it into the ST-GNNs to obtain the forecasting result. The
detail procedure are listed as follow:
\begin{alignat}{2}
	& \text{Generating Weight:} &&\ \left\{w_k \right\}_{k=1}^K  = f(x_{test})\nonumber, \\
	&\text{Graphons Mixup:}&&\ \mathcal{P}_{mix} = \sum\limits_{k=1}^{K}\frac{e^{w_k}}{\sum\nolimits_{j}e^{w_j}} \mathcal{P}_k, \nonumber \\
	&\text{Feeding into ST-GNNs:}&&\quad  \hat{y}_{test} = g(x_{test}, G), \ G \sim \mathcal{P}_{mix}, \nonumber 
\end{alignat}




\section{Experiment}

\noindent\textbf{Datasets.} We conducted experiments on four datasets, namely PEMS03, PEMS04, PEMS07, and PEMS08, evaluating their performance in both in-distribution and out-of-distribution scenarios.  To evaluate model performance over time, we further propose four variants under OOD
scenarios: PEMS03-2019, PEMS04-2019, PEMS07-2018, and PEMS08-2017, which align with existing standards \cite{song2020spatial, guo2019attention}, using the same sensors to capture traffic data from different years.  
The experimental setup involved predicting the 12-step future based on the 12-step historical data \cite{li2018diffusion, wu2019graph}. To further validate the model's generalization capability, we collected real-time traffic data provided by the Traffic Management Center (TMC)\footnote{\url{https://www.nyc.gov/html/dot/html/motorist/atis.shtml}} of the New York City Department of Transportation. The model was trained using Speed data from March to May 2019, and tested on a multi-year dataset spanning from 2020 to 2022. The dataset partitioning follows the same approach as used in PEMS.
The ratio of these three subsets is approximately 6:2:2.

\begin{table*}[t]
	\caption{Forecasting performance comparison of different approaches on PEMS03, PEMS04, PEMS07 and PEMS08 datasets. We use arrows to indicate that the model is trained on in-distribution data and tested on out-of-distribution data.}\label{tab:ours_forecast}
	\centering
	\renewcommand{\arraystretch}{1.3}
	\begin{sc}
		\resizebox{\textwidth}{!}{
			\begin{tabular}{lccccccccccccc}
				\toprule
				\multirow{2}{*}{\textbf{Model}} & \multirow{2}{*}{\textbf{Params}}& \multicolumn{3}{c}{\textbf{PEMS03-2018}} & \multicolumn{3}{c}{\textbf{PEMS03-2018 $\rightarrow$ 2019}} & \multicolumn{3}{c}{\textbf{PEMS04-2018}} & \multicolumn{3}{c}{\textbf{PEMS04-2018 $\rightarrow$ 2019}}
				\\
				\cmidrule(lr){3-5}\cmidrule(lr){6-8} \cmidrule(lr){9-11} \cmidrule(lr){12-14}
				& & MAE & RMSE & MAPE & MAE & RMSE & MAPE & MAE & RMSE & MAPE & MAE & RMSE & MAPE\\
				\hline
				CauSTG &247K &18.48 &29.31 &25.78\% &26.82 &54.83 &15.53\% &24.07 &37.85 &18.55\% &28.78 &47.77 &22.40\% \\
				CaST &285K  &17.86 &28.05 &25.83\% &24.23 &52.42 &14.73\% &23.89 &36.37 &17.37\% &27.96 &46.47 &21.35\% \\
				STONE &741K & 17.29 & 27.09 & 24.40\% & 23.52 & 50.83 & 12.18\% & 22.50 & 35.34 & 16.56\% & 26.37 & 46.23 & 80.24\% \\ 
				\midrule
				GWNET &303K & 14.37&23.01&14.52\% & 16.14& 25.32& 16.81\% &18.53&30.72&12.62\% & 33.16& 51.74& 25.57\% \\
				\rowcolor{Gray}  +Expert &324K   & \bf14.25& \bf22.87& \bf14.38\% & \bf 16.02& \bf 24.86& \bf 15.78\% &\bf18.41&\bf 30.50\bf&12.69\% & \bf 26.01& \bf 41.81& \bf 19.60\% \\
				\hline
				TrendGCN &457K & 14.69&23.73&15.02\% & 16.33& 25.42& 16.63\% &18.91&33.84&12.90\% & 38.15& 59.62& 30.53\% \\
				\rowcolor{Gray} +Expert & 473K &\bf 14.54&\bf 23.53& \bf14.99\% & \bf 15.63& \bf 24.92& \bf 15.90\% &\bf18.82&\bf 33.63&\bf 12.85\% & \bf 32.40& \bf 50.06& \bf 27.04\% \\
				\hline
				AGCRN &757K & 15.41&25.62&14.49\% & 18.87& 29.92& 19.44\% &19.24&32.50&13.14\% & 45.64& 69.37& 35.63\% \\
				\rowcolor{Gray} +Expert & 776K &\bf 15.31&\bf 25.45&\bf 14.16\% & \bf 18.06& \bf 28.31& \bf 18.21\% &\bf19.05&\bf 32.36& \bf 13.07\% & \bf 32.74& \bf 49.30& \bf 26.31\% \\
				\hline
				MTGNN & 621K&  15.26&24.96&16.71\% & 18.88& 29.59& 19.71\% &19.93&33.51&13.08\% & 48.08& 74.09& 40.29\% \\
				\rowcolor{Gray} +Expert & 641K &\bf 15.23&\bf24.82&\bf16.40\% & \bf18.17& \bf 28.89& \bf 18.92\% &\bf19.81&\bf 33.37&\bf13.01\% & \bf 32.83& \bf 50.05& \bf 26.80\% \\
				\hline
				STAEformer & 1.41M & 14.53&23.62&15.44\% & 16.83& 26.24& 19.07\% &18.53&30.62&12.44\% & 35.07& 53.83& 28.07\% \\
				\rowcolor{Gray} +Expert & 1.42M  & \bf 14.43&\bf23.51&\bf 15.37\% & \bf14.17&\bf23.60&\bf10.06\% &\bf18.45&\bf30.58&\bf 12.18\% & \bf27.09&\bf41.91&\bf20.31\% \\
				\bottomrule
				\multirow{2}{*}{\textbf{Model}} & \multirow{2}{*}{\textbf{Params}}& \multicolumn{3}{c}{\textbf{PEMS07-2017}} & \multicolumn{3}{c}{\textbf{PEMS07-2017 $\rightarrow$ 2018}} & \multicolumn{3}{c}{\textbf{PEMS08-2016}} & \multicolumn{3}{c}{\textbf{PEMS08-2016 $\rightarrow$ 2017}}
				\\
				\cmidrule(lr){3-5}\cmidrule(lr){6-8} \cmidrule(lr){9-11} \cmidrule(lr){12-14}
				& & MAE & RMSE & MAPE & MAE & RMSE & MAPE & MAE & RMSE & MAPE & MAE & RMSE & MAPE\\
				\hline
				CauSTG &247K &26.16 &41.67 &15.16\% &29.02 &44.31 &25.36\% &19.48 &29.31 &18.78\% &23.48 &34.31 &36.78\% \\
				CaST &285K &25.87 &40.25 &14.61\% &28.63 &42.39 &24.40\% &18.86 &28.05 &18.83\% &22.86 &33.05 &35.83\% \\
				STONE &741K & 24.95 & 38.88 & 13.70\% & 27.05 & 41.49 & 22.10\% & 17.81 & 27.71 & 17.28\% & 17.82 & 28.72 & 34.26\% \\ 
				\midrule
				GWNET &303K &20.10&33.12&10.96\% & 33.52& 50.07& 19.87\% &14.55&24.74&9.00\% & 34.86& 47.39& 28.46\% \\
				\rowcolor{Gray}  +Expert &324K   &\bf20.05&\bf32.99&\bf9.49\% & \bf26.50& \bf 40.41& \bf 14.91\% &\bf14.26&\bf24.42&\bf8.96\% & \bf 17.14& \bf 28.52& \bf 11.03\% \\
				\hline
				TrendGCN &457K &19.79&33.22&9.92\% & 36.91& 52.77& 43.50\% &14.48&24.73&8.90\% & 31.08& 46.00& 26.24\% \\
				\rowcolor{Gray} +Expert & 473K &\bf19.64&\bf 33.03&\bf8.39\% & \bf 26.62& \bf 41.32& \bf 23.94\% &\bf14.35&\bf 24.56&\bf8.73\% & \bf 27.56& \bf 41.79& \bf 29.84\% \\
				\hline
				AGCRN &757K &21.69&34.30&10.16\% & 56.17& 78.82& 108.25\% &14.90&25.38&8.61\% & 48.57& 68.24& 50.44\% \\
				\rowcolor{Gray} +Expert & 776K &\bf21.52&\bf 34.08&\bf9.55\% & \bf 38.26& \bf 53.09& \bf 58.72\% &\bf14.79&\bf 25.07&\bf8.28\% & \bf 30.24& \bf 43.93& \bf 32.80\% \\
				\hline
				MTGNN & 621K&21.92&34.04&10.03\% & 52.36& 73.42& 73.31\% &14.70&24.86&8.55\% & 47.00& 66.59& 49.38\% \\
				\rowcolor{Gray} +Expert & 641K &\bf21.87&\bf33.82&\bf9.76\% & \bf 40.60& \bf 56.78& \bf 55.40\% &\bf14.53&\bf24.57&\bf8.41\% & \bf 26.88& \bf 40.29& \bf 25.13\% \\
				\hline
				STAEformer & 1.41M &19.31&32.88&9.33\% & 35.07& 53.83& 28.07\% &13.44&23.30&9.07\% & 32.96& 49.66& 28.85\% \\
				\rowcolor{Gray} +Expert & 1.42M  &\bf 19.14&\bf 32.80&\bf 8.31\% & \bf24.16&\bf23.29&\bf19.37\% &\bf13.36&\bf23.05&\bf8.78\% & \bf 17.03 &\bf 27.24 &\bf 11.89\% \\
				\bottomrule
			\end{tabular}
		}
	\end{sc}
\end{table*}

\noindent\textbf{Experiment Settings}
Our experiments were conducted on a GPU server with eight GeForce GTX 3090 graphics cards, utilizing the PyTorch 2.0.3 framework. We standardized raw data through z-score normalization \cite{cheadle2003analysis}. Training was terminated early if the validation error stabilized within 15-20 epochs or showed no improvement after 200 epochs, with the best model retained based on validation data \cite{luo2023dynamic}. We faithfully followed the model parameters and settings from the original paper, and also performed multiple rounds of parameter tuning to optimize the results. To ensure a comprehensive evaluation of the state's traffic conditions, we selected the same sensors as the current benchmark. Following the methodology outlined in \cite{guo2019attention}, we chronologically split the data into train, validation, and test sets, maintaining a ratio of 6:2:2 across all datasets.  Model performance was evaluated using Mask-Based Root Mean Square Error (RMSE), Mean Absolute Error (MAE), and Mean Absolute Percentage Error (MAPE), excluding zero values as they represent noisy data \cite{jiang2021dl}.  Notably, we exclude zero values (representing noisy data) from these metrics \cite{jiang2021dl}. The raw data undergoes standardization using Z-Score \cite{cheadle2003analysis}. In case the validation error converges within 15-20 epochs or ceases after 100 epochs, the training process concludes prematurely, saving the best model based on the validation data \cite{luo2023dynamic}.

\begin{table*}[t]
	\caption{To evaluate the robustness of the expert beyond PEMS system, we compared New York speed data over three years.}\label{tab:nyc}
	\centering
	\renewcommand{\arraystretch}{1.3}
	\begin{sc}
		\resizebox{\textwidth}{!}{
			\begin{tabular}{lccccccccccccc}
				\toprule
				\multirow{2}{*}{\textbf{Model}} & \multirow{2}{*}{\textbf{Params}}& \multicolumn{3}{c}{\textbf{SpeedNYC-2019}} & \multicolumn{3}{c}{\textbf{SpeedNYC-2019 $\rightarrow$ 2020}} & \multicolumn{3}{c}{\textbf{SpeedNYC-2019 $\rightarrow$ 2021}} &  \multicolumn{3}{c} {\textbf{SpeedNYC-2019 $\rightarrow$ 2022}}
				\\
				\cmidrule(lr){3-5}\cmidrule(lr){6-8} \cmidrule(lr){9-11} \cmidrule(lr){12-14}
				& & MAE & RMSE & MAPE & MAE & RMSE & MAPE & MAE & RMSE & MAPE & MAE & RMSE & MAPE\\
				\hline
				CauSTG &247K &4.85 &7.80 &26.20\% &6.90 &10.10 &26.80\% &7.55 &10.85        &44.50\% &6.10 &9.00 &36.20\% \\
				CaST &285K &4.80 &7.75 &25.90\% &6.85 &10.05 &26.50\% &7.50 &10.80 &44.00\% &6.05 &8.95 &36.00\% \\
				STONE &741K &4.75 &7.70 &25.60\% &6.80 &10.00 &26.20\% &7.45 &10.75 &43.50\% &6.00 &8.90 &35.80\% \\
				\midrule
				GWNET &303K & 4.77 & 7.66 & 25.54\% & 6.79 & 9.95 & 26.11\% & 7.41 & 10.68 & 43.56\% & 5.99 & 8.84 & 35.55\% \\
				\rowcolor{Gray} +Expert &324K &\bf4.72 &\bf7.60 &\bf25.30\% &\bf6.50 &\bf9.70 &\bf25.00\% &\bf7.10 &\bf10.30 &\bf41.00\% &\bf5.70 &\bf8.50 &\bf33.00\% \\
				\hline
				TrendGCN &457K & 4.72 & 7.71 & 26.26\% & 6.43 & 9.91 & 26.41\% & 7.11 & 10.27 & 40.70\% & 6.06 & 8.96 & 37.60\% \\
				\rowcolor{Gray} +Expert &478K &\bf4.67 &\bf7.65 &\bf26.00\% &\bf6.20 &\bf9.70 &\bf25.50\% &\bf6.90 &\bf10.00 &\bf39.00\% &\bf5.80 &\bf8.70 &\bf35.00\% \\
				\hline
				AGCRN &757K & 4.89 & 7.90 & 26.89\% & 7.01 & 10.16 & 26.49\% & 7.84 & 11.17 & 44.84\% & 6.97 & 10.17 & 43.71\% \\
				\rowcolor{Gray} +Expert &778K &\bf4.84 &\bf7.84 &\bf26.60\% &\bf6.70 &\bf9.90 &\bf25.50\% &\bf7.50 &\bf10.80 &\bf42.00\% &\bf6.60 &\bf9.80 &\bf41.00\% \\
				\hline
				MTGNN &621K & 4.72 & 7.70 & 25.93\% & 6.54 & 10.01 & 27.34\% & 8.15 & 11.50 & 47.21\% & 6.67 & 9.80 & 43.57\% \\
				\rowcolor{Gray} +Expert &642K &\bf4.67 &\bf7.64 &\bf25.70\% &\bf6.30 &\bf9.80 &\bf26.50\% &\bf7.80 &\bf11.20 &\bf45.00\% &\bf6.40 &\bf9.50 &\bf41.00\% \\
				\hline
				STAEformer&1.41M & 4.70 & 7.68 & 24.96\% & 6.15 & 9.46 & 24.84\% & 7.16 & 10.42 & 41.43\% & 6.02 & 8.89 & 34.53\% \\
				\rowcolor{Gray} +Expert &1.42M &\bf4.65 &\bf7.62 &\bf24.70\% &\bf5.90 &\bf9.20 &\bf24.00\% &\bf6.90 &\bf10.10 &\bf39.00\% &\bf5.80 &\bf8.60 &\bf32.00\% \\
				\bottomrule
			\end{tabular}
		}
	\end{sc}
\end{table*}

\begin{table*}[h]
	\caption{ Ablation study of our method performing in GWNET.} \label{tab:abla}
	\begin{sc}
		\begin{subtable}
			\centering
			\small
			\resizebox{!}{0.096\linewidth}{
				\begin{tabular}{llccc}
					\hline Dataset & Architecture & MAE & RMSE & MAPE \\
					\hline \multirowcell{4}{ \bf PEMS04\\ \bf 2018}& Ours  &  \bf 19.21& \bf 30.40&\bf 13.29\% \\
					& W/o episodic   & 20.55 & 31.74 & 15.87\%\\
					& W/o weight    & 22.67 & 33.01 & 18.14\%\\
					& W/o mixup  & 19.84&  30.55& 13.19\% \\
					\hline
					\multirowcell{4}{ \bf PEMS08\\ \bf 2016}& Ours  &  \bf 14.26& \bf 23.42&\bf 9.06\% \\
					& W/o episodic  & 15.11 & 23.44 & 10.03\% \\
					& W/o weight  & 18.81 & 25.98 & 14.47\%  \\
					& W/o mixup   & 14.32 & 24.52 & 10.17\%   \\
					\hline
			\end{tabular}}
		\end{subtable}
		\begin{subtable}
			\centering
			\small
			\resizebox{!}{0.096\linewidth}{
				\begin{tabular}{llccc}
					\hline Dataset & Architecture & MAE & RMSE & MAPE \\
					\hline \multirowcell{4}{ \bf PEMS04\\ \bf 2018 $\rightarrow$ 2019}& Ours  &  \bf 26.01& \bf 41.81& \bf 19.60\% \\
					& W/o episodic   & 27.66 & 43.71 & 21.84\%\\
					& W/o weight    & 29.86 & 48.04 & 24.17\%\\
					& W/o mixup  & 31.22 & 50.04 & 25.09\%  \\
					\hline
					\multirowcell{4}{ \bf PEMS08\\ \bf 2016$\rightarrow$ 2017}& Ours  &  \bf 17.14& \bf 28.52&\bf  11.03\% \\
					& W/o episodic  & 22.34 & 35.04 & 14.81\% \\
					& W/o weight  & 24.12 & 38.84 & 17.77\%  \\
					& W/o mixup   & 27.51 & 42.57 & 20.51\%   \\
					\hline
			\end{tabular}}
		\end{subtable}
	\end{sc}
\end{table*}

\noindent\textbf{Baseline Methods.}
We adopt the following representative traffic forecasting baselines.  LSTM \cite{hochreiter1997long} is a temporal-only deep model that does not consider the spatial correlations. Taking advantage of the advancements in GNNs \cite{defferrard2016convolutional,kipf2017semi}, sequential models have been integrated with GNNs to effectively model traffic data. During the period from 2018 to 2024, we specifically select RNN-based methods such as and DGCRN \cite{li2023dynamic}, AGCRN \cite{bai2020adaptive}, TrendGCN \cite{jiang2023enhancing}, TCN-based methods like STGCN \cite{yu2018spatio}, MTGNN~\cite{wu2020connecting} and GWNET \cite{wu2019graph}, as well as attention-based methods ASTGCN \cite{guo2019attention} and STTN \cite{xu2020spatial}. Moreover, we incorporate five recent representative methods, which reflect the recent research directions in the field. STGODE \cite{fang2021spatial} leverages neural ordinary differential equations to effectively model the continuous changes of traffic signals. DSTAGNN \cite{lan2022dstagnn}, DGCRN \cite{li2023dynamic},  D$^2$STGNN \cite{shao2022decoupled} and STAEformer \cite{liu2023spatio}  specifically consider the dynamic characteristics of correlations among sensors on traffic networks.  We also compare the recent spatiotemporal OOD model CauSTG \cite{zhou2023maintaining}, STONE \cite{wang2024stone}, CaST \cite{xia2024deciphering}, which are designed to capture invariant relationships for OOD problem.

\noindent\textbf{Comparing with Spatiotemporal OOD Method.}\label{sec:ood_baseline} First, we conducted a thorough evaluation of spatiotemporal OOD  approaches. A shared characteristic among STONE, CaST, and CauSTG is their significantly inferior performance compared to current mainstream methods on in-distribution data. Nevertheless, these methods exhibit improved performance in OOD scenarios, although they still suffer from noticeable performance degradation. Notably, in the PEMS03 dataset, as previously analyzed, no substantial distributional shift is observed, resulting in strong overall network performance. However, all OOD methods experienced considerable performance degradation in this scenario, which leads us to question whether they genuinely capture invariant features. In fact, we argue that spatiotemporal invariance is inherently difficult to define. For example, predicting when future traffic disruptions might occur due to construction or the development of new commercial centers or hospitals is inherently uncertain. We hypothesize that the failure of current methods can largely be attributed to this unpredictability. In contrast, the modeling approach in our expert framework adopts a different perspective. Rather than attempting to identify invariant characteristics, our method emphasizes exposing the model to distributional shifts during training, thereby enhancing its robustness against spatiotemporal changes. Consequently, our approach demonstrates strong performance in both in-distribution and out-of-distribution settings.

\noindent\textbf{Expert Performance on PEMS Dataset.}\label{sec:out} \tableref{tab:ours_forecast} compares the forecasting performance of various models, including GWNET, TrendGCN, AGCRN, MTGNN, and STAEformer, with and without our method on the PEMS03, PEMS04, PEMS07, PEMS08, and cross-year datasets. The arrow indicates models trained on in-distribution data and tested on out-of-distribution data. STAEformer, similar to other models, incorporates Adaptive Embedding; however, it distinguishes itself by embedding similarity computation directly within its attention mechanism, foregoing the explicit construction of a graph from the embeddings. In line with proposed framework, we train $K$ Adaptive Embedding Experts through episodic learning, subsequently aggregating their outputs during inference to adapt to previously unseen distributions.

Our results show that incorporating our method improves performance in relation shift scenarios while maintaining performance in the original distribution. The STAEformer model shows the most significant improvements on the PEMS07 and PEMS08 datasets, with enhancements of 56.72\% and 45.15\%, respectively. Similarly, MTGNN demonstrates remarkable improvements on the PEMS04 and PEMS08 datasets, achieving 32.44\% and 39.50\% improvements, respectively. These results highlight the models' ability to adapt to different spatiotemporal dynamics across varying datasets.
Notably, the minimal improvement on PEMS03 is likely due to its slight spatial shift (\figref{fig:kendall_03}). In contrast, PEMS08 shows superior performance, partly due to its smaller graph size (170 nodes vs. 307 and 883 nodes in PEMS04 and PEMS07), which simplifies learning. 
These results underscore the effectiveness of our method in improving model performance in relation shift scenarios and highlight the importance of choosing appropriate models and methods based on the specific dataset and task requirements.

\begin{table*}[t]
	\caption{Forecasting performance comparison of different approaches on PEMS series and SpeedNYC datasets. The arrow is used to illustrate the transfer learning results.}\label{tab:forecast_result}
	\centering
	\begin{sc}
		\resizebox{\textwidth}{!}{
			\begin{tabular}{lcccccccccccc}
				\toprule
				\multirow{2}{*}{\textbf{Model}}& \multicolumn{3}{c}{\textbf{PEMS03-2019}} & \multicolumn{3}{c}{\textbf{PEMS03-2018 $\rightarrow$ 2019}}
				& \multicolumn{3}{c}{\textbf{PEMS04-2019}} & \multicolumn{3}{c}{\textbf{PEMS04-2018 $\rightarrow$ 2019}}
				\\
				\cmidrule(lr){2-4}\cmidrule(lr){5-7} \cmidrule(lr){8-10} \cmidrule(lr){11-13}
				& MAE & RMSE & MAPE & MAE & RMSE & MAPE & MAE & RMSE & MAPE & MAE & RMSE & MAPE\\
				\midrule
				ASTGCN~  	& 16.84& 25.97& 18.93\%& 22.37& 34.70& 21.66\% & 22.39& 37.00& 14.95\%& 49.68& 72.48& 35.90\%  \\
				STGODE~		& 16.21& 26.16& 16.09\%& 18.00& 28.29& 19.12\% & 20.70& 34.52& 14.94\%& 36.35& 56.12& 30.23\%  \\
				DSTAGNN~		& 16.10& 25.84& 15.26\%& 20.94& 31.85& 20.16\% & 19.95& 33.04& 13.88\%& 41.86& 60.02& 39.96\%  \\
				MTGNN~		& 15.26& 24.96& 16.71\%& 18.88& 29.59& 19.71\% & 19.93& 33.51& 13.08\%& 48.08& 74.09& 40.29\%  \\
				AGCRN~		& 15.41& 25.62& 14.49\%& 18.87& 29.92& 19.44\% & 19.24& 32.50& 13.14\%& 45.64& 69.37& 35.63\%  \\
				STGCN~			& 16.29& 27.31& 14.95\%& 20.79& 32.71& 24.29\% & 19.75& 32.09& 13.93\%& 53.18& 83.50& 46.94\%  \\
				\rowcolor{Gray} LSTM~ 	& 16.24& 25.23& 15.84\%& 16.50& 25.61& \bf 16.06\% & 23.42& 38.78& 16.54\%&\bf 24.26&\bf 39.17&\bf 17.50\%  \\
				STTN~			& 15.24& 24.05& 14.54\%& 16.93& 26.12& 17.05\% & 20.34& 34.20& 15.64\%& 37.30& 57.95& 29.84\%  \\
				GWNET~			& \bf 14.37&\bf 23.01& 14.52\%& 16.14& 25.32& 16.81\% & 18.53& 30.72& 12.62\%& 33.16& 51.74& 25.57\%  \\
				DGCRN~			& 14.61& 23.36& \bf 14.51\%& 16.58& 25.70& 17.12\% & 18.86& 30.91& 12.67\%& 37.08& 56.07& 28.93\%  \\
				D$^2$STGNN  & 14.88& 24.25& 14.19\%& 16.95& 26.75& 19.57\% & 18.28& 30.12& 12.68\%& 38.17& 61.24& 30.10\%  \\
				TrendGCN 		& 14.69& 23.73& 15.02\%& \bf 16.33&\bf 25.42& 16.63\% & 18.91& 33.84&   12.90\%& 38.15& 59.62& 30.53\%  \\
				STAEformer  & 14.53& 23.62& 15.44\%& 16.83& 26.24& 19.07\%& \bf 18.53&\bf  30.62&\bf 12.44\%& 35.07& 53.83& 28.07\% \\
				\toprule
				\multirow{2}{*}{\textbf{Model}}& \multicolumn{3}{c}{\textbf{PEMS07-2018}} & \multicolumn{3}{c}{\textbf{PEMS07-2017 $\rightarrow$ 2018}}
				& \multicolumn{3}{c}{\textbf{PEMS08-2017}} & \multicolumn{3}{c}{\textbf{PEMS08-2016 $\rightarrow$ 2017}}
				\\
				\cmidrule(lr){2-4}\cmidrule(lr){5-7} \cmidrule(lr){8-10} \cmidrule(lr){11-13}
				& MAE & RMSE & MAPE & MAE & RMSE & MAPE & MAE & RMSE & MAPE & MAE & RMSE & MAPE\\
				\midrule
				ASTGCN~  	& 22.61& 38.02& 13.94\%& 58.88& 84.12& 49.04\%   & 15.31& 27.75& 9.82\%& 32.50& 47.49& 27.52\% \\
				STGODE~		& 20.26& 34.35& 10.62\%& 46.32& 66.01& 41.64\%   & 15.21& 25.57& 8.59\%& 39.73& 59.04& 33.62\% \\
				DSTAGNN~		& 21.51& 34.50& 12.41\%& 52.36& 69.74& 86.04\%   & 15.20& 27.58& 9.67\%& 37.82& 52.73& 40.83\% \\
				MTGNN~		& 21.92& 34.04&  10.03\%& 52.36& 73.42& 73.31\%   & 14.70& 24.86& \bf 8.55\%& 47.00& 66.59& 49.38\% \\
				AGCRN~		& 21.69& 34.30& 10.16\%& 56.17& 78.82& 108.25\%  & 14.90& 25.38& 8.61\%& 48.57& 68.24& 50.44\% \\
				STGCN~			& 22.85& 35.54& 11.02\%& 88.18& 124.68& 187.79\% & 15.71& 26.35& 9.25\%& 67.47& 99.00& 68.35\% \\
				\rowcolor{Gray} LSTM~  	& 24.96& 38.53& 10.98\%& \bf 25.32& \bf 38.89&\bf 22.06\%   & 16.52& 27.82& 9.88\%& \bf 16.87& \bf 28.44&\bf 9.38\%  \\
				STTN~			& 21.83& 33.91& 10.09\%& 39.27& 55.98& 34.47\%   & 15.05& 25.06& 8.87\%& 42.17& 61.04& 28.78\% \\
				GWNET~			& 20.10& 33.12& 10.96\%& 33.52& 50.07& 19.87\%   & 14.55& 24.74& 9.00\%& 34.86& 47.39& 28.46\% \\
				DGCRN~			& 20.21& 34.19& 12.45\%& 37.70& 53.33& 41.85\%   & 15.33& 25.74& 8.82\%& 30.14& 44.71& 29.05\% \\
				D$^2$STGNN  & 19.49& 32.96& 9.38\%& 42.92& 61.68& 37.81\%   &  14.37& 24.51& 9.97\%& 33.45& 49.53& 35.34\% \\
				TrendGCN 		&  19.79& 33.22& 9.92\%& 36.91& 52.77& 43.50\%   & 14.48& 24.73& 8.90\%& 31.08& 46.00& 26.24\% \\
				STAEformer 		& \bf 19.31&\bf 32.88&\bf 9.33\%& 35.07& 53.83& \bf 28.07\%&\bf 13.44&\bf 23.30& 9.07\%& 32.96& 49.66& 28.85\%\\
				\toprule
				\multirow{2}{*}{\textbf{Model}}& \multicolumn{3}{c}{\textbf{SpeedNYC-2019}} & \multicolumn{3}{c}{\textbf{SpeedNYC-2019 $\rightarrow$ 2020}} & \multicolumn{3}{c}{\textbf{SpeedNYC-2019 $\rightarrow$ 2021}} &  \multicolumn{3}{c} {\textbf{SpeedNYC-2019 $\rightarrow$ 2022}}
				\\
				\cmidrule(lr){2-4}\cmidrule(lr){5-7} \cmidrule(lr){8-10} \cmidrule(lr){11-13}
				& MAE & RMSE & MAPE & MAE & RMSE & MAPE & MAE & RMSE & MAPE & MAE & RMSE & MAPE\\
				\midrule
				AGCRN & 4.89 & 7.90 & 26.89\% & 7.01 & 10.16 & 26.49\% & 7.84 & 11.17 & 44.84\% & 6.97 & 10.17 & 43.71\% \\
				D$^{2}$STGNN & \bf 4.56 & \bf 7.54 & 23.76\% & 6.39 & 9.89 & 25.42\% & 7.20 & 10.60 & 42.10\% & 6.03 & 9.13 & 36.24\% \\
				TrendGCN & 4.72 & 7.71 & 26.26\% & 6.43 & 9.91 & 26.41\% & 7.11 & 10.27 & 40.70\% & 6.06 & 8.96 & 37.60\% \\
				GWNET & 4.77 & 7.66 & 25.54\% & 6.79 & 9.95 & 26.11\% & 7.41 & 10.68 & 43.56\% & 5.99 & 8.84 & 35.55\% \\
				\rowcolor{Gray}  LSTM & 5.19 & 8.29 & 26.45\% & \bf 4.50 & \bf 7.79 & \bf 18.50\% &\bf 6.07 &\bf 9.80 & \bf35.78\% & \bf 5.54 &\bf 8.63 &\bf 28.96\% \\
				MTGNN & 4.72 & 7.70 & 25.93\% & 6.54 & 10.01 & 27.34\% & 8.15 & 11.50 & 47.21\% & 6.67 & 9.80 & 43.57\% \\
				STAEformer & 4.70 & 7.68 & 24.96\% & 6.15 & 9.46 & 24.84\% & 7.16 & 10.42 & 41.43\% & 6.02 & 8.89 & 34.53\% \\
				STGCN & 4.68 & 7.63 & 24.20\% & 7.46 & 10.66 & 27.19\% & 8.31 & 11.67 & 49.33\% & 6.59 & 9.74 & 41.08\% \\
				STGODE & 4.82 & 7.66 & 26.38\% & 6.62 & 9.79 & 24.65\% & 8.32 & 11.80 & 52.44\% & 6.11 & 8.97 & 37.32\% \\
				STTN & 4.93 & 8.05 & 27.45\% & 6.69 & 10.04 & 27.30\% & 8.41 & 12.06 & 50.06\% & 7.13 & 10.35 & 47.10\% \\
				\bottomrule
		\end{tabular}}
	\end{sc}
	\vspace{-10pt}
\end{table*}

\noindent\textbf{Performance Comparison beyond PEMS System.} To compare the performance of the expert framework on non-PEMS datasets, we present results for multi-year dataset scenarios in \tableref{tab:nyc}. For more detailed results, please refer to  \tableref{tab:forecast_result}. The LSTM results align with observations on PEMS data, where significant shifts in spatial information were noted. This further supports our hypothesis that traffic data, influenced by factors such as traffic regulations and road width, is predominantly shaped by spatial drift. As shown in \tableref{tab:nyc}, We observe that all models show improvements across all cross-year scenarios. The GWNET owns best performance, with the most significant improvement of 3.85\% in the 2019 to 2022. Meanwhile, STAEformer performs optimally in the 2019 to 2020 scenario, achieving an improvement of 2.75\%. 

\begin{figure*}[t]
	\centering
	\includegraphics[width=1\linewidth]{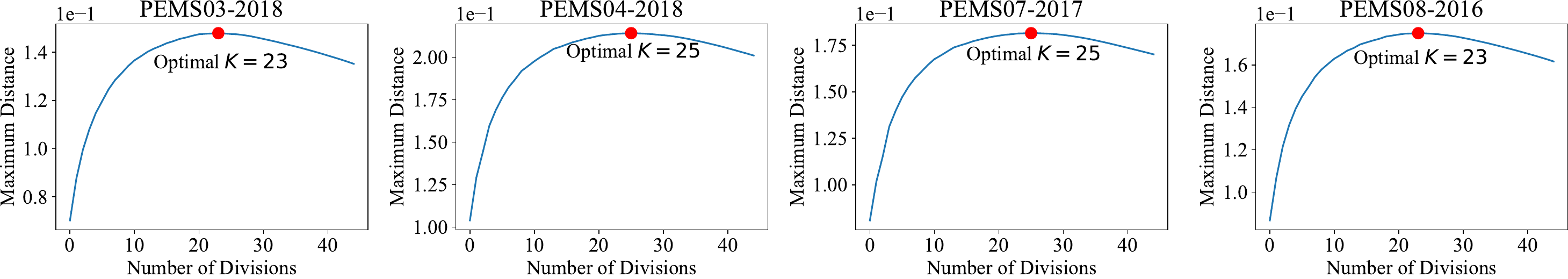}
	\includegraphics[width=1\linewidth]{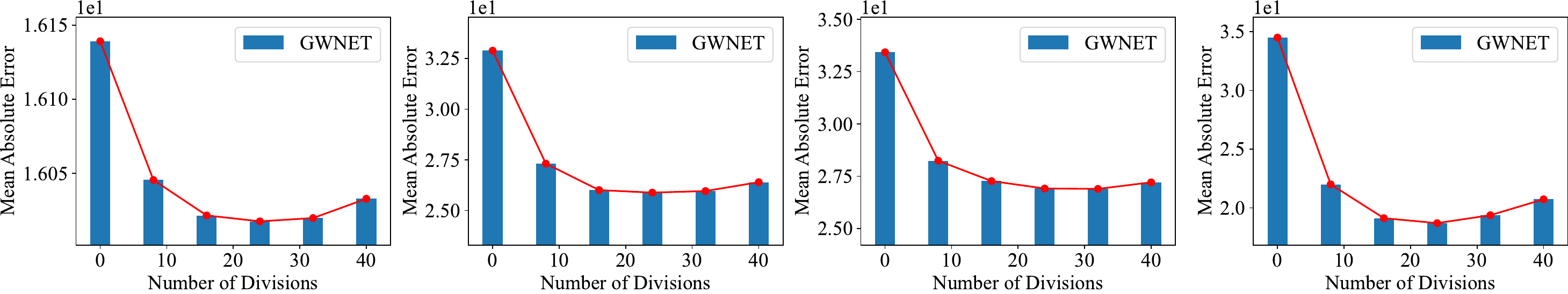}
	\includegraphics[width=1\linewidth]{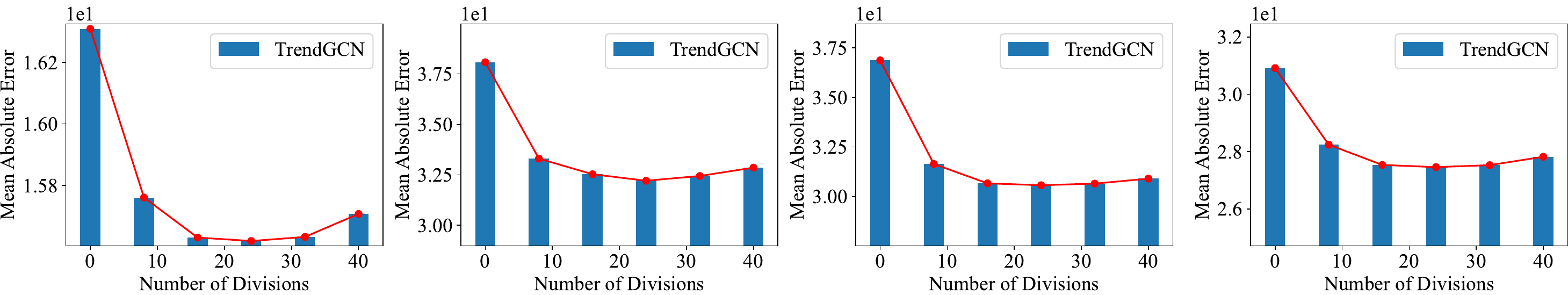}
	\includegraphics[width=1\linewidth]{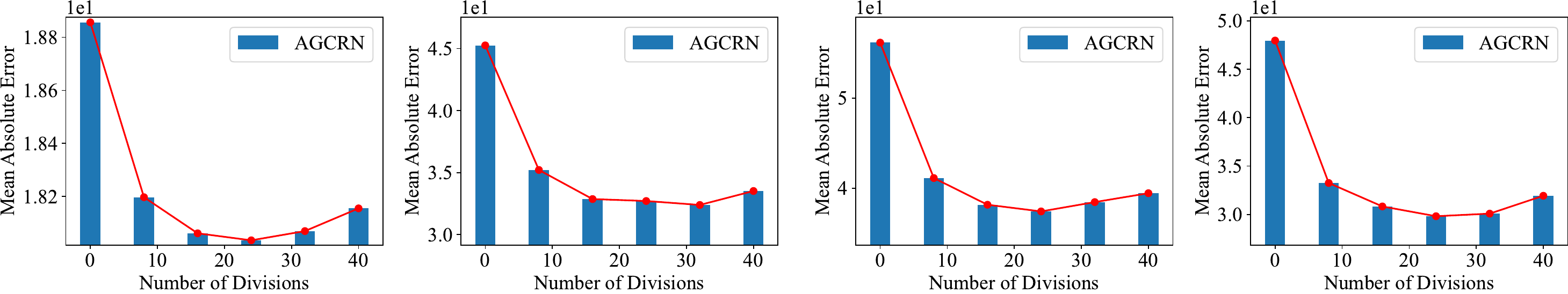}
	\includegraphics[width=1\linewidth]{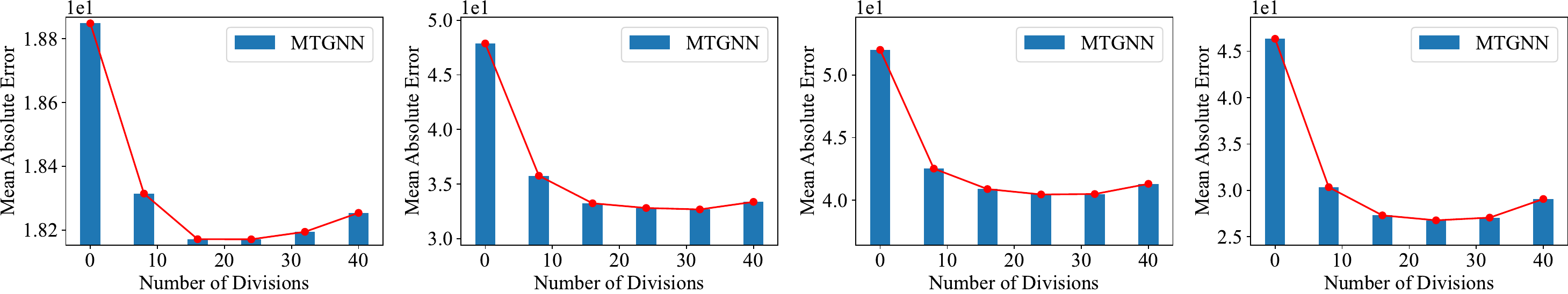}
	\caption{Searching for the optimal number of divisions is depicted for each dataset in the upper figures. The corresponding performance of GWNET, TrendGCN, AGCRN, and MTGNN  under different numbers of divisions for each dataset is illustrated at the bottom of the figure.}\label{fig:optimal_k}
	
\end{figure*}

\noindent\textbf{Ablation Study.}\label{sec:Ablation}
To assess the relative importance of each component, we conducted ablation studies on the PEMS04 and PEMS08 datasets, evaluating both in-distribution and out-of-distribution performance. The following scenarios were explored:
1) without episodic training: instead of aggregating multiple experts via episodic training, we utilized exponential moving averages to collect statistical information across different partitions during training. At inference time, experts were aggregated based on a weighted arithmetic mean distance between sample features and the previously collected training statistics.
2) without mixup: experts were aggregated with equal probabilities, applying uniform weighting across all experts.
3 ) without weighting: test samples were categorized using the same partitioning scheme employed during training, and the corresponding expert was used for inference.
For in-distribution performance (PEMS03-2018 and PEMS08-2016), our approach exhibited comparable results to the scenario without mixup, indicating that the stability of graph relationships was maintained. The scenario without episodic training outperformed the scenario without weighting, which mixed all graphons equally.
Regarding out-of-distribution performance, removing the mixup module significantly impacted multi-step forecasting for both PEMS04 and PEMS08, suggesting that expert graphons without mixup are less effective. However, even without mixup, performance was superior to using a single expert (GWNET). The equal weighting scenario outperformed the scenario without mixup on both datasets.

\noindent\textbf{Optimal Divisions.}\label{sec:division} To address scalability and efficiency issues in large-scale data, we implemented a dynamic programming algorithm and set a minimum sequence length for each expert (\(\alpha_1 = 6\)). We conducted a grid search on each training dataset to determine the optimal number of partitions. Given a specified \(K\), we identified partitions that maximize the distribution distance based on \eqnref{equ:obj}. 
Under the given constraints, we found that the maximum distance partition occurs near in different dataset. To verify whether the maximum distance partition aligns with the model's generalization performance, we compared it to the results from a grid search method used to define the number of partitions. As shown in Figure \ref{fig:optimal_k}, the maximum distance significantly increases around \([23, 25]\) before subsequently declining. Empirically, the algorithm tends to combine early morning peaks, late evening peaks, and midnight into a single group.

\noindent\textbf{Relation between  MSGD  and Generalization Ability.}\label{sec:relation} we present a performance comparison of GWNet, TrendGCN, AGCRN, and MTGNN under various divisions in Figure \ref{fig:optimal_k} to showcase the effectiveness of our searching algorithm described in Equation \eqnref{equ:obj} on Maximum Spatiotemporal Graph Division. An interesting phenomenon we observed is the consistency between the model's generalization ability and the maximum partition distance. Specifically, the greater the distance between partitioned data, the stronger the model's generalization capability, which aligns with our previous hypothesis regarding the diversification of trained experts. Based on this finding, extensive grid searches are unnecessary; we can simply substitute the formula to determine the optimal number of partitions \(K\) before training the model.
As mentioned earlier, division 0 represents the vanilla model without ST-expert graphons. The results clearly demonstrate substantial improvements across all divisions, emphasizing the effectiveness of our method. Notably, each sub-figure exhibits an obvious turning point, underscoring that blindly increasing the number of experts is not a reasonable choice. This is due to the increased memory burden and the potential for insufficient training samples for each expert.

\noindent\textbf{Performance Details}
In \tableref{tab:forecast_result}, we evaluated the performance of various ST-GNNs on traffic forecasting tasks using the PEMS03, PEMS04, PEMS07, and PEMS08 datasets, including out-of-distribution scenarios. The models were assessed for both same-year forecasting and cross-year transfer learning scenarios. The results highlighted the notable strength of LSTMs in handling cross-year data variations, where they consistently outperformed several advanced ST-GNNs. Our findings aligned closely with the data analysis results shown in \figref{fig:motivation}, which revealed that while GNN models effectively capture spatiotemporal relationships in traffic data, as evidenced by their strong performance in same-year forecasts, they generally underperform in cross-year transfer learning tasks compared to LSTM models. This underperformance in transfer learning scenarios suggests a potential limitation of ST-GNNs in learning diverse and robust graph relationships, despite their demonstrated prowess in spatial relationship modeling. These insights underscore the importance of selecting suitable models based on the specific requirements and temporal dynamics of the traffic forecasting task at hand.



	
	

\section{Conclusion}\label{sec:con}
In this paper, we investigate the performance of state-of-the-art models using extended traffic benchmarks and find a significant decline in their traffic forecasting accuracy over time. Our careful analysis attributes this decline to the models' inability to adapt to unseen spatial dependencies. To address this challenge, we propose a novel MoE framework that learns a set of graphons during training and adaptively mixes them to tackle the spatial distribution shift problem during testing. We extend this concept to the Transformer architecture, achieving substantial improvements in performance.  Our method is simple yet effective and can be seamlessly integrated into any ST-model, significantly outperforming current state-of-the-art models in handling spatial dynamic issues.  While the proposed expert framework targets enhanced adaptability to unobserved spatial dependencies, the dynamic changes in spatial dependence may surpass the current model's capacity to capture fully. Real-world spatial dependencies are influenced by a myriad of factors such as irregular urban development, shifts in transportation policies, significant events, and more.


\bibliographystyle{IEEEtran}
\bibliography{reference}
\input{people}

\end{document}

%% file: people.tex
\begin{IEEEbiography}[{\includegraphics[width=1in,height=1.25in,clip,keepaspectratio]{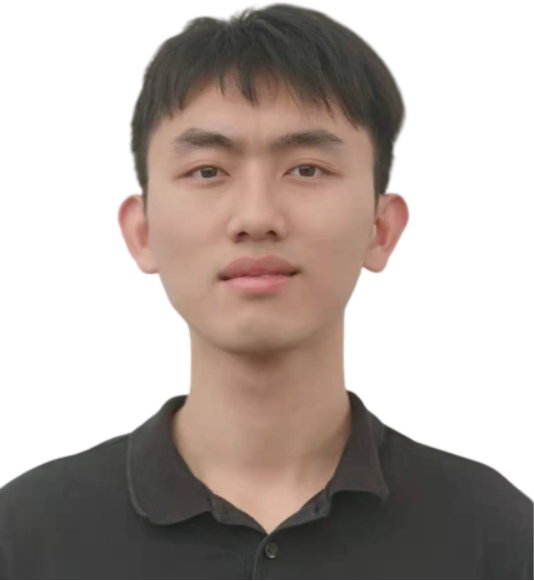}}]{Hongjun Wang} is working toward the PhD degree in the Department of Mechano-Informatics at The University of Tokyo. He received his M.S. degree in computer science and technology from Southern University of Science and Technology, China. He received his B.E. degree from the Nanjing University of Posts and Telecommunications, China, in 2019. His research interests are broadly in machine learning, with a focus on urban computing, explainable AI, data mining, and data visualization.
\end{IEEEbiography}
\vspace{1ex}
\begin{IEEEbiography}[{\includegraphics[width=1in,height=1.25in,clip,keepaspectratio]{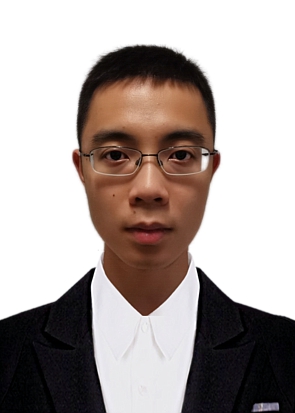}}]{Jiyuan Chen} is working towards his PhD degree at The Hong Kong Polytechnic University. He received his B.S. degree in Computer Science and Technology from Southern University of Science and Technology, China. His major research fields include artificial intelligence, deep learning, urban computing, and data mining.
\end{IEEEbiography}
\vspace{1ex}
\begin{IEEEbiography}[{\includegraphics[width=1in,height=1.25in,clip,keepaspectratio]{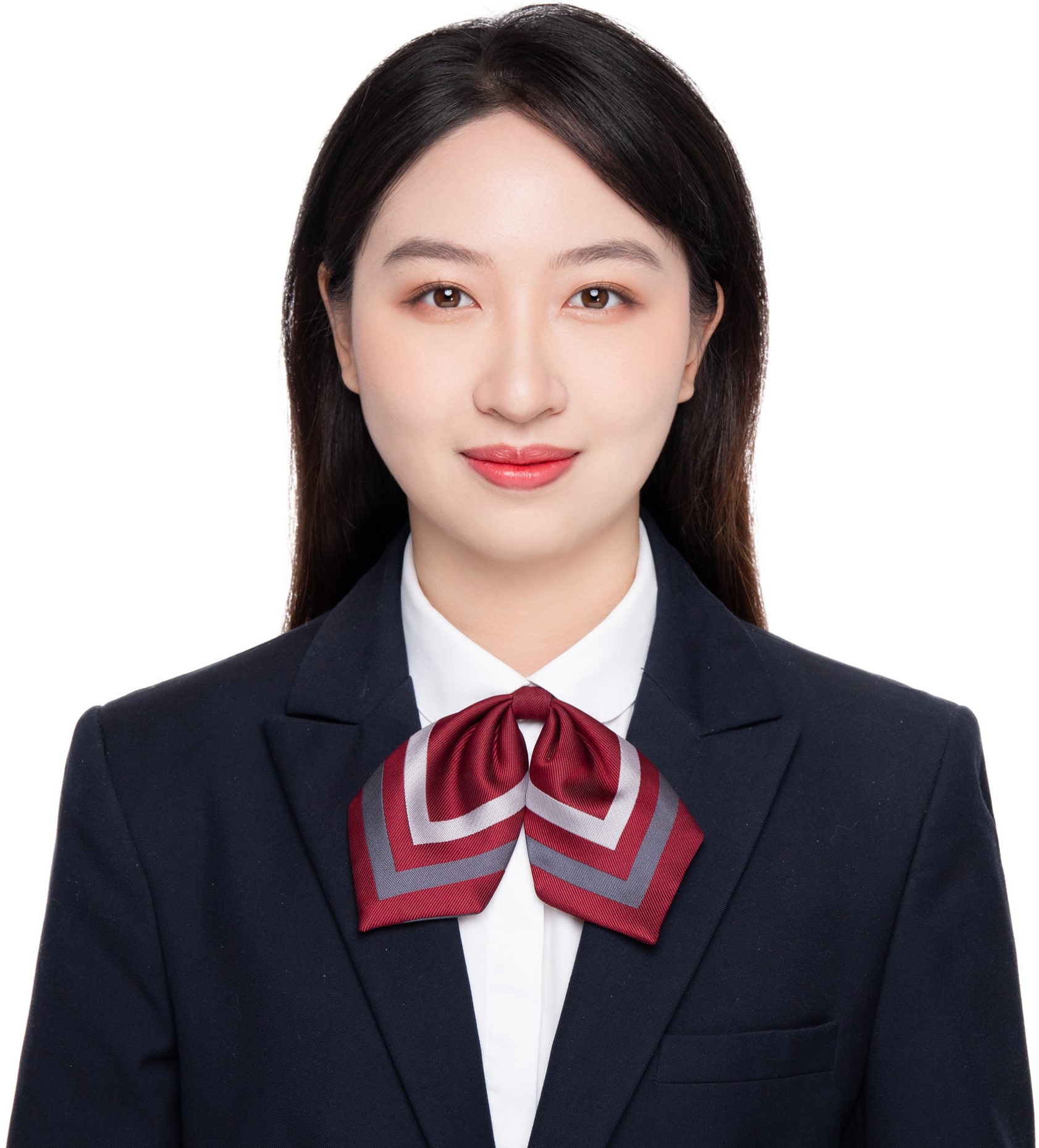}}]{Tong Pan} received a B.S. degree in Physics from East China Normal University, China, in 2019, and a Ph.D. degree in Physics from the Chinese University of Hong Kong, China, in 2024. From 2024, she has been a postdoctal researcher at Southern University of Science and Technology, China. Her research interests include data analysis, machine learning and AI for science. 
\end{IEEEbiography}
\vspace{1ex}
\begin{IEEEbiography}[{\includegraphics[width=1in,height=1.25in,clip,keepaspectratio]{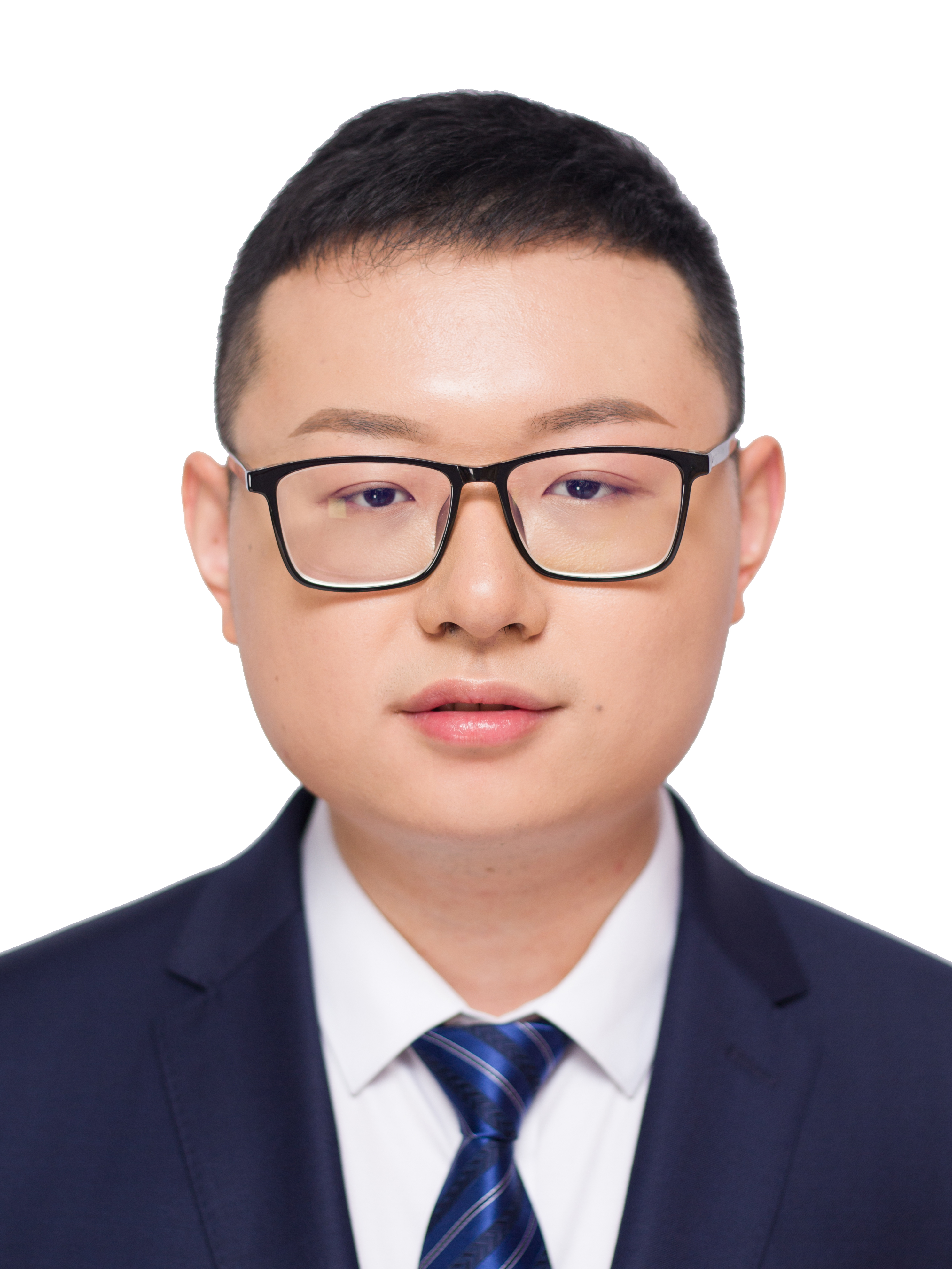}}]{Zheng Dong}  received his B.E. degree in computer science and technology from Southern University of Science and Technology (SUSTech) in 2022. He is currently persuing a M.S. degree in the Department of Computer Science and Engineering, SUSTech. His research interests include deep learning and spatio-temporal data mining.
\end{IEEEbiography}

\vspace{1ex}
\begin{IEEEbiography}[{\includegraphics[width=1in,height=1.25in,clip,keepaspectratio]{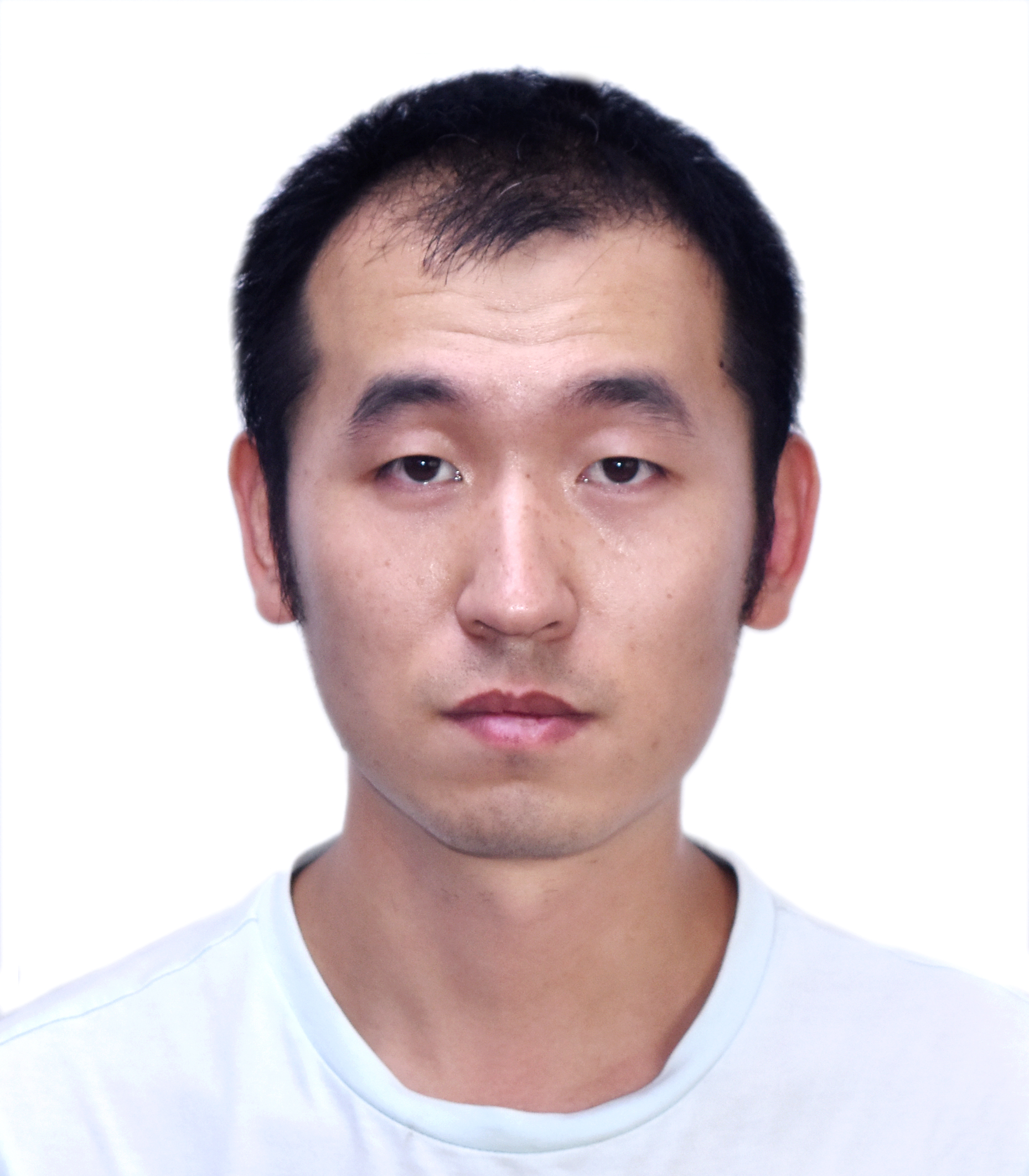}}]{Lingyu Zhang}   joined Baidu in 2012 as a search strategy algorithm research and development engineer. He joined Didi in 2013 and served as senior algorithm engineer, technical director of taxi strategy algorithm direction, and technical expert of strategy model department. Currently a researcher at Didi AI Labs, he used machine learning and big data technology to design and lead the implementation of multiple company-level intelligent system engines during his work at Didi, such as the order distribution system based on combination optimization, and the capacity based on density clustering and global optimization. Scheduling engine, traffic guidance and personalized recommendation engine, "Guess where you are going" personalized destination recommendation system, etc. Participated in the company's dozens of international and domestic core technology innovation patent research and development, application, good at using mathematical modeling, business model abstraction, machine learning, etc. to solve practical business problems. He has won honorary titles such as Beijing Invention and Innovation Patent Gold Award and QCon Star Lecturer, and his research results have been included in top international conferences related to artificial intelligence and data mining such as KDD, SIGIR, AAAI, and CIKM.
\end{IEEEbiography}
\vspace{1ex}
\begin{IEEEbiography}[{\includegraphics[width=1in,height=1.25in,clip,keepaspectratio]{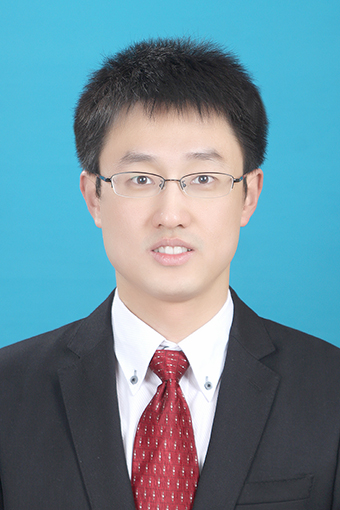}}]{Renhe Jiang} received a B.S. degree in software engineering from the Dalian University of Technology, China, in 2012, a M.S. degree in information science from Nagoya University, Japan, in 2015, and a Ph.D. degree in civil engineering from The University of Tokyo, Japan, in 2019. From 2019, he has been an Assistant Professor at the Information Technology Center, The University of Tokyo. His research interests include ubiquitous computing, deep learning, and spatio-temporal data analysis.
\end{IEEEbiography}
\vspace{1ex}
\begin{IEEEbiography}[{\includegraphics[width=1in,height=1.25in,clip,keepaspectratio]{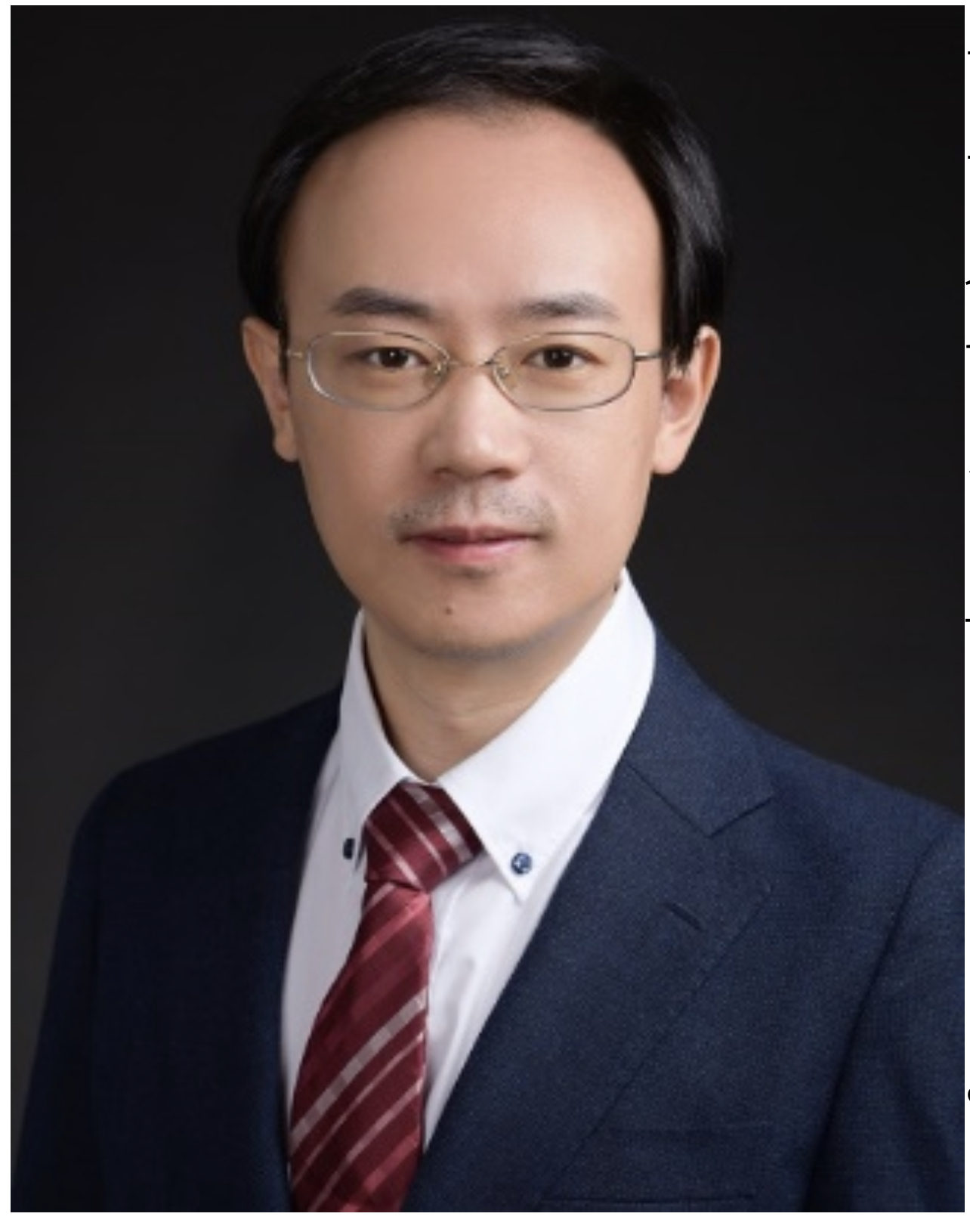}}]{Prof. Xuan Song}  received the Ph.D. degree in signal and information processing from Peking University in 2010. In 2017, he was selected as an Excellent Young Researcher of Japan MEXT. In the past ten years, he led and participated in many important projects as a principal investigator or primary actor in Japan, such as the DIAS/GRENE Grant of MEXT, Japan; Japan/US Big Data and Disaster Project of JST, Japan; Young Scientists Grant and Scientific Research Grant of MEXT, Japan; Research Grant of MLIT, Japan; CORE Project of Microsoft; Grant of JR EAST Company and Hitachi Company, Japan. He served as Associate Editor, Guest Editor, Area Chair, Program Committee Member or reviewer for many famous journals and top-tier conferences, such as IMWUT, IEEE Transactions on Multimedia, WWW Journal, Big Data Journal, ISTC, MIPR, ACM TIST, IEEE TKDE, UbiComp, ICCV, CVPR, ICRA and etc.
\end{IEEEbiography}